\documentclass[accepted, specialissue]{melba}

\usepackage{graphicx}
\usepackage{amsfonts}
\usepackage{amsmath}
\usepackage{booktabs}
\usepackage{color}
\usepackage{comment}
\usepackage{epstopdf}
\usepackage{float}
\usepackage[edges]{forest}
\usepackage{multirow}
\usepackage{multicol}
\usepackage{subcaption}
\usepackage{textgreek}
\usepackage{tikz}
\usepackage{xurl}
\usepackage[table]{xcolor}

\newcommand{\MDD}{M100}
\newcommand{\MBD}{F25M75}
\newcommand{\ESD}{F50M50}
\newcommand{\FBD}{F75M25}
\newcommand{\FDD}{F100}

\definecolor{myred}{rgb}{.8,.0,.0}

\newcommand{\rebuttal}[1]{\textcolor{black}{#1}}

\melbaid{2026:011}  
\doi{10.59275/j.melba.2026-4156}
\melbaauthors{Raumanns, Schouten, Pluim and Cheplygina}  
\email{ralf.raumanns@fontys.nl}
\volume{2026}
\firstpageno{200}  
\melbayear{2026}  
\datesubmitted{2025-04-21}  
\datepublished{2026-05-29}  

\melbaspecialissue{Special issue on Fairness of AI in Medical Imaging (FAIMI)}
\melbaspecialissueeditors{Veronika Cheplygina, Aasa Feragen, Andrew King, Ben Glocker, Enzo Ferrante, Eike Petersen, Esther Puyol-Antón, Melanie Ganz-Benjaminsen}

\ShortHeadings{Demographic Bias in Skin Lesions}{Raumanns, Schouten, Pluim and Cheplygina}

\title{Effect of Demographic Bias on Skin Lesion Classification}

\author{
	\firstname Ralf \surname Raumanns\aff{1,3,4}\orcid{0000-0002-7055-8136},
	\name Gerard Schouten\aff{2}\orcid{0000-0001-7042-2143},
    \name Veronika Cheplygina\aff{4}\orcid{0000-0003-0176-9324},
    \name Josien P.W. Pluim\aff{3}\orcid{0000-0001-7327-9178}}

\affiliations{
	\num 1 \addr Fontys University of Applied Science, Venlo, The Netherlands  \\
	\num 2 \addr Fontys University of Applied Science, Eindhoven, The Netherlands \\
	\num 3 \addr Eindhoven University of Technology, Eindhoven, The Netherlands \\
    \num 4 \addr IT University of Copenhagen, Denmark
}

\abstract{
    The influence of bias in datasets on the fairness of model predictions is a topic of ongoing research in various fields. In this study, we evaluate the performance of skin lesion classification using ResNet-based convolutional models, focusing on the impact of demographic bias in training data, particularly variations in patient sex and age. We use a linear programming method to generate datasets with controlled demographic characteristics, allowing systematic investigation of bias effects. Three distinct learning strategies are evaluated: a single-task model, a reinforcing multi-task model, and an adversarial learning scheme.

\rebuttal{Our sex-based analysis indicates that sex-specific training datasets optimise model performance. Notably, including male patients in the training data improved performance for the male subgroup, even in female-majority cases. Reinforcing and adversarial learning schemes narrowed or eliminated bias gaps in balanced and female-majority datasets. However, these strategies proved less effective in male-majority settings, where models continued to perform better for males than females. The two learning schemes showed marginal bias reduction compared to the baseline model in predominantly male patient populations.}

\rebuttal{Age-based analysis demonstrates comparable baseline performance across the three model approaches, with performance declining across age categories. Younger groups consistently achieve the highest performance, regardless of training data distribution. Although balanced training yields optimal results for the youngest age category, performance decreases in older categories.}

\rebuttal{We find that sex biases arise mainly from data imbalances, while age biases consistently favour younger groups regardless of distribution. These distinct mechanisms require targeted mitigation strategies. Our work aims to advance equitable AI in medical imaging by addressing these specific sources of disparity.}

\rebuttal{Additionally, cross-dataset validation on two external datasets revealed that domain shifts notably affect performance and demographic bias patterns.}
 
The source code and models are available on GitHub: \\\url{https://github.com/raumannsr/demographic-fairness-extended}.}
    

\begin{document}

\twocolumn[\maketitle]

\section{Introduction}\label{Introduction}
Deep learning has shown many successes in the diagnosis of medical images, as demonstrated by several studies ( \cite{Saha2024-nq,Esteva2017-hn,Ehteshami_Bejnordi2017-gr}\rebuttal{)}, but despite the high overall performance, models can be biased against patients from different demographic groups, a concern highlighted in recent work (\cite{abbasi2020risk,larrazabal2020gender,gichoya2022ai}). Bias and fairness have therefore become central research topics in medical imaging, with studies focusing, for example, on skin lesions (\cite{abbasi2020risk,Groh2021-vm}), chest radiographs (\cite{larrazabal2020gender}) and brain magnetic resonance imaging \rebuttal{(}\cite{petersen2022feature}). Sensitive attributes commonly examined are age, sex, or race. For the classification of skin lesions, the Fitzpatrick skin type is often studied (\cite{Seth2024-ai,Bencevic2024-pg,Groh2021-vm,Wu2022-hl}).

While deep learning models continue to advance diagnostic capabilities, their fairness remains a significant concern because model performance is fundamentally tied to the quality and representativeness of the training data, as well as the model's ability to mitigate any bias embedded in the training dataset.

Although bias and fairness in AI for medical imaging have gained attention, prior studies have often examined individual demographic factors in isolation, typically within a single imaging modality \rebuttal{or without systematic control over data distributions. A comprehensive evaluation comparing how these demographic attributes, when systematically skewed, influence model performance across different learning strategies (single-task, reinforcing, and adversarial) is lacking.} Moreover, the relative effectiveness of debiasing approaches across specific demographic subgroups, particularly under extreme distributional imbalances, remains unexplored. \rebuttal{Additionally, the utility of auxiliary demographic prediction heads as fairness indicators has not been systematically assessed.}

In this paper, we define dataset bias (also known as representation bias) strictly as demographic bias, meaning any systematic imbalance in age, sex, or other protected attributes within the training set. Such imbalances lead to unbalanced learning and performance gaps between subgroups. We examine both demographic and model bias, measuring how controlled skews in the training data affect performance, and testing multi-task learning strategies designed to mitigate model bias. Using a balanced test set, we quantify the degree to which demographic bias propagates to model bias and identify the most effective approaches for equitable skin-lesion classification across age and sex groups.

This manuscript substantially extends our FAIMI 2024 workshop paper (\cite{Raumanns2025-nt}). The workshop paper evaluated five distributions of male/female patients (sex demographics) with three learning strategies (one single-task and two multi-task models). \rebuttal{The evaluation} focused on overall and subgroup-specific performance to assess whether training data distribution biases manifested in results when tested on a balanced test set.


Extending our FAIMI 2024 workshop paper, we present the following contributions: 
\begin{enumerate} 
    \item We extend our linear programming (LP) method to control age subgroups in addition to sex, introducing five age groups and three skewed age distributions. 
    \item We systematically evaluate two bias mitigation strategies (reinforcing multi-task and adversarial) across various age and sex subgroups. By presenting both overall and subgroup-specific metrics, we determine how each strategy performs under different conditions. This includes two new sex-distribution scenarios, namely predominantly male and predominantly female patients, which enable a more granular evaluation of the models.
    \item Beyond the internal hold-out validation, we extend our external validation from the prior study. We introduce a new dermatoscopic skin-lesion dataset in this work. Alongside the retained smartphone dataset, the datasets facilitate testing across diverse geographical regions, acquisition methods, and demographic groups.
    \item We analyse the auxiliary age-prediction head to assess its utility as a fairness indicator. 
\end{enumerate}

\section{Related work}\label{RelatedWorks}
    We revisit prior studies on demographic bias and fairness in medical imaging, highlighting how earlier work has examined demographic disparities, bias mitigation techniques such as multi-task and adversarial learning, and the limitations that motivate our more systematic analysis.

\paragraph{Understanding representation bias}
    Demographic bias in medical imaging, referring to performance disparities across protected attributes (such as biological sex, race, age, and skin tone), has been extensively studied, revealing how these imbalances can cause unfair or discriminatory outcomes in healthcare. Glocker et al. showed that a widely used chest radiography foundation model actually encodes protected attributes, like biological sex and race, leading to statistically significant performance gaps across those subpopulations (\cite{Glocker2023-db}). Vaidya et al. reported that deep learning pathology models exhibit racial bias, as demonstrated on large publicly available cancer imaging datasets (\cite{Vaidya2024-xz}).

    Demographic bias in machine learning manifests in various forms, with representation bias being particularly significant in healthcare. Representation bias occurs when certain demographic groups are underrepresented in training data, leading to reduced model performance for these groups (\cite{larrazabal2020gender}). This differs from bias caused by inherent anatomical or physiological differences between groups, though these can contribute to representation bias when they affect data collection, for example, clinical protocols that exclude pregnant patients for safety reasons (\cite{Seyyed-Kalantari2021-dv}).

    Sies et al. assessed a market-approved skin cancer CNN and documented a male predominance in the training data. Despite this imbalance, performance on a balanced test set showed no statistically significant sex-related disparity, suggesting the extensive training set mitigated the imbalance effect (\cite{Sies2022-dp}). Conversely, even with deliberately balanced datasets, intrinsic anatomical differences can still generate bias. Klingenberg et al. demonstrated this by showing that a CNN trained on a sex-balanced MRI cohort for Alzheimer's detection performed markedly better in female patients than males, underscoring that demographic bias can arise from physiological factors rather than merely data imbalance (\cite{Klingenberg2023-rk}).

    Understanding how representation bias influences model performance is essential for building fair systems. By identifying underrepresented populations or those with the poorest performance, targeted corrections can be applied to the dataset. Moreover, understanding these effects provides insights for designing future datasets, allowing researchers to avoid similar problems early on.

\paragraph{Role of demographics}
    The role of demographics in medical AI is multifaceted. Some demographic variations reflect genuine biological differences that models should take into account; for example, patient characteristics such as age and sex significantly influence the predictive precision of health markers, such as blood pressure, in retinal image analysis (\cite{Gerrits2021-yo}). Deep learning models can extract demographic characteristics, such as sex and age, directly from medical images, such as chest X‑rays, with high accuracy \rebuttal{(\cite{Gichoya2022-nn};\cite{Jones2025-zn})}. This capability offers applications in forensic investigations, aiding identification and uncovering novel anatomical landmarks for sex and age determination (\cite{Yi2021-nm}).
    
    However, it is crucial to distinguish these valid demographic correlations from problematic representation bias, which often relates to data collection practices rather than physiological differences. \rebuttal{Our research addresses this by deliberately building datasets with specific demographic imbalances, helping us determine whether performance disparities are due to true physiological factors or simply to data collection.}

\paragraph{Addressing bias}

    \rebuttal{Research on fairness usually involves baseline studies that demonstrate bias between groups and/or suggest methods to enhance fairness. These approaches mainly tackle representation bias through sampling or weighting strategies during training (\cite{Groh2021-vm}). Alternatively, they implement architectural techniques that prevent models from depending on sensitive attributes, such as adversarial learning (\cite{abbasi2020risk}). For example, Yang et al. developed an adversarial framework to mitigate biases arising from hospital location and patient ethnicity (\cite{Yang2023-cw}). Wu et al. introduced FairPrune, which trims parameters based on their importance to both privileged and unprivileged groups (\cite{Wu2022-hl}). Other methods focus on data augmentation. Stanley et al. proposed a synthetic bias framework for brain MRI. They showed that simple sample reweighting effectively reduces hidden biases (\cite{Stanley2024-yw}). Ktena et al. demonstrated that diffusion-generated synthetic images improve fairness across histopathology, chest X-ray, and dermatology datasets (\cite{Ktena2024-bt}).}
    
    \rebuttal{Commonly used datasets for studying demographic bias in skin lesion classification include the ISIC skin lesion datasets (\cite{Gutman2016-lf, Codella2017-rd, Codella2019-cn, Tschandl2018-sz, Combalia2019-jj, Veronica2021patient}) and Fitzpatrick-17K (\cite{Groh2021-vm, Groh2022-ja}).}
    \rebuttal{However, researchers typically rely on pre-provided data splits or stratify by a single demographic attribute (e.g., male vs female). Crucially, these methods often fail to control for the interplay between attributes, treating sex and age as independent variables rather than managing their joint distribution. Our linear programming approach, however, explicitly enforces constraints on both sex and age simultaneously, ensuring that specific subgroups (such as older males or younger females) are accurately represented according to the desired ratios.}

\paragraph{\rebuttal{Bias mitigation approaches}}
    Our current study builds on two crucial insights from medical imaging: multi-task learning and shortcut learning (\cite{Geirhos2020-te,Nauta2021-fe}).     
    \rebuttal{The reinforcing model uses multi-task learning, which trains multiple related tasks simultaneously, aiming to enhance the model's generalisability while also reducing bias in two ways. Firstly, when the same hidden layers support multiple related tasks (\cite{ruder2017overview}), the network must learn features that work across various contexts. Benefiting one task over another is not the goal, as this would diminish performance on other tasks. Joint training of the tasks improves generalisability, as each task regularises the others (\cite{Caruana1993}). Secondly, training a multi-task model requires more diverse data than a single-task model; in addition to medical image data, demographics are also included in the training. This learning approach exposes the model to a broader range of data, which may help reduce the influence of patterns that are prominent only in a subset of the data. Having more data lets multi-task models build stronger, more general features that work across several tasks, helping prevent overfitting (\cite{Zhang2022}). This suggests that incorporating an auxiliary task, specifically addressing potential bias factors such as age and sex, alongside the primary binary classification (malignant or not), could help mitigate bias. }

    \rebuttal{In addition to the standard multi-task approach, our study also employs an adversarial model approach, a special variant of multi-task learning. As previously demonstrated (\cite{Adeli2021-da,abbasi2020risk}), this strategy reduces output bias to some degree through adversarial training. The model aims to minimise bias by decreasing the mutual information between learned features and the protected attribute, employing a negative-squared Pearson correlation loss for age and binary cross-entropy for sex.}

    \rebuttal{We use the auxiliary head's performance as a diagnostic tool for the reinforcing model. Moderate to high accuracy confirms that the demographic signal has been learned and that regularisation is active, serving as a direct indicator that the debiasing mechanism is functioning properly.}
    
    Studies have explored different approaches to handling demographic attributes in model training. Some use demographics within multi-task learning settings (\cite{liu2019-gz}), where attributes reinforce diagnosis during optimisation. This contrasts with more recent adversarial strategies (\cite{Adeli2021-da,abbasi2020risk}) that specifically aim to reduce representation bias by preventing models from predicting sensitive attributes. Additionally, representation bias can be confounded by correlations between demographics and imaging characteristics, leading to shortcut learning. These characteristics include variations in imaging devices (such as different scanner types or image acquisition protocols) and technical artefacts such as surgical markers or medical instruments \rebuttal{(}\cite{willemink2020preparing,jimenez2023detecting,gichoya2022ai,Bissoto2020-hn}). For example, Bevan and Atapour-Abarghouei specifically demonstrated how these technical artefacts can introduce bias in the classification of skin lesions, developing methods to identify and mitigate their impact (\cite{Bevan2023-kg}). In such cases, addressing representation bias requires considering multiple confounding factors, as balancing data for one demographic attribute may leave other sources of bias unaddressed.
    
    We aim to comprehensively evaluate AI‑based skin‑lesion classification models across demographic groups, with a particular focus on identifying and mitigating representation bias. Whereas Sies et al. examined a market-approved CNN in its uncontrolled training set and considered only sex bias (\cite{Sies2022-dp}), we deliberately construct subsets with exact sex and age ratios to study how \rebuttal{the combination of} demographic skews affect performance.


\section{Methods}\label{Methods}
    To evaluate the impact of demographic imbalance in training data on skin-lesion classification, we conducted two parallel experiments: one manipulating the distribution of patient sex and another modifying the age distribution. In the sex-based analysis, we created datasets with different male-to-female ratios. In the age-based analysis, we built datasets with skewed age profiles, favouring younger, older, or balanced age groups, \rebuttal{while keeping a 1:1 sex ratio}. Both analyses followed the same methodological pipeline, using linear programming, with the only difference being the demographic attribute constrained during dataset creation. We first describe the data collection and preprocessing steps, then outline the model architectures and evaluation methods.

\subsection{Data}
    \begin{table*}[hbt!]
\centering
\caption{\rebuttal{Overview of the curated skin-lesion datasets used in this study.}}

\newcolumntype{L}{>{\color{black}}c}
\newcolumntype{C}{>{\color{black}}c}

\small
\begin{tabular}{LLCCCCC}
Dataset & Modality & Total & Malignant & Benign& Age & Geographic\\
 &  & samples & (female/male) & (female/male) & info & origin\\
\hline
ISIC & Dermoscopic & 35,884 & 7,074 (3,012/4,062) & 28,810 (13,207/15,603) & Yes & Global \\
PAD-UFES-20 & Smartphone & 1,179 & 833 (401/432) & 346 (198/148) & Yes & Brazil\\
DERM7PT & Dermoscopic & 1,011 & 294 (160/134) & 718 (362/355) & No & Italy\\
\end{tabular}
\label{tab:overview_skin_lesion_datasets}
\end{table*}

\begin{figure*}[t]
        \centering
        \begin{minipage}{0.3\textwidth}
            \centering
            \begin{subfigure}[b]{0.45\textwidth}
                \includegraphics[width=\textwidth]{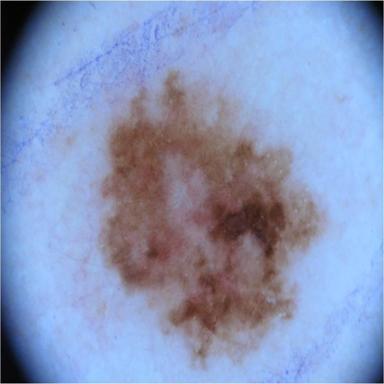}
                \caption{F. malignant}
            \end{subfigure}
            \hfill
            \begin{subfigure}[b]{0.45\textwidth}
                \includegraphics[width=\textwidth]{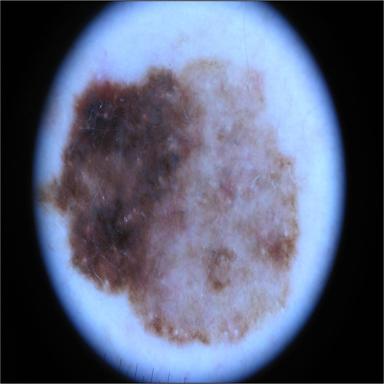}
                \caption{M. malignant}
            \end{subfigure}
            \vfill
            \begin{subfigure}[b]{0.45\textwidth}
                \includegraphics[width=\textwidth]{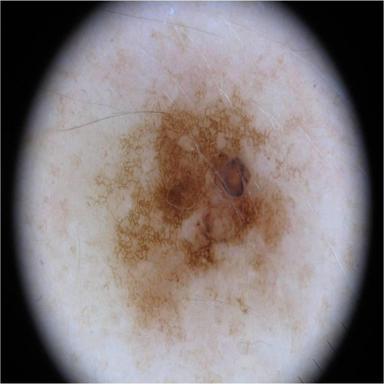}
                \caption{F. benign}
            \end{subfigure}
            \hfill
            \begin{subfigure}[b]{0.45\textwidth}
                \includegraphics[width=\textwidth]{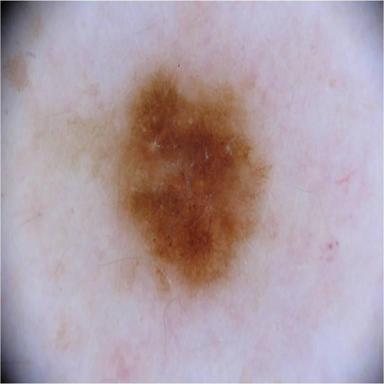}
                \caption{M. benign}
            \end{subfigure}
        \end{minipage}%
        \hfill
        \begin{minipage}{0.3\textwidth}
            \centering
            \begin{subfigure}[b]{0.45\textwidth}
                \includegraphics[width=\textwidth]{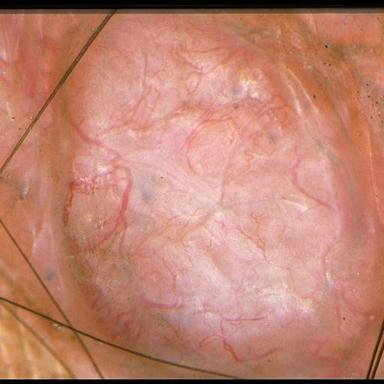}
                \caption{F. malignant}
            \end{subfigure}
            \hfill
            \begin{subfigure}[b]{0.45\textwidth}
                \includegraphics[width=\textwidth]{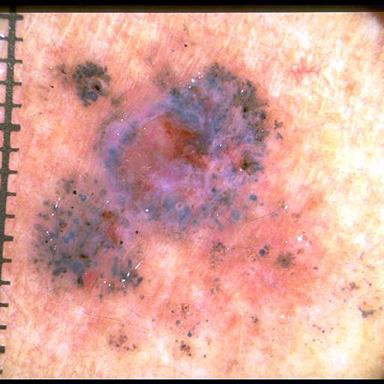}
                \caption{M. malignant}
            \end{subfigure}
            \vfill
            \begin{subfigure}[b]{0.45\textwidth}
                \includegraphics[width=\textwidth]{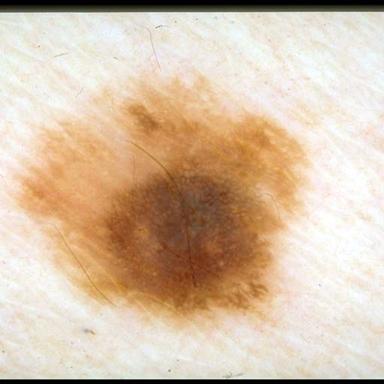}
                \caption{F. benign}
            \end{subfigure}
            \hfill
            \begin{subfigure}[b]{0.45\textwidth}
                \includegraphics[width=\textwidth]{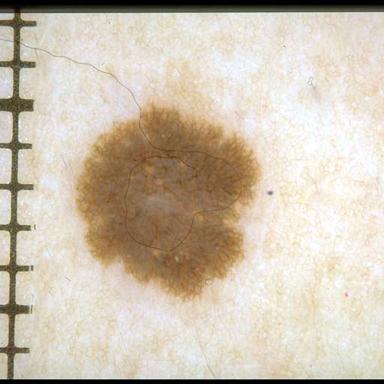}
                \caption{M. benign}
            \end{subfigure}
        \end{minipage}%
        \hfill
        \begin{minipage}{0.3\textwidth}
            \centering
            \begin{subfigure}[b]{0.45\textwidth}
                \includegraphics[width=\textwidth]{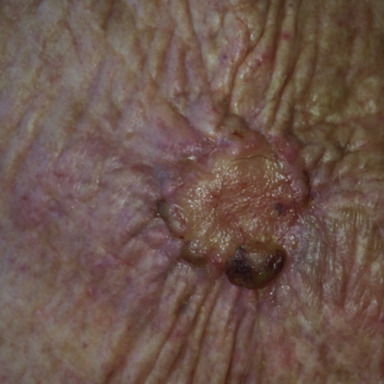}
                \caption{F. Malignant}
            \end{subfigure}
            \hfill
            \begin{subfigure}[b]{0.45\textwidth}
                \includegraphics[width=\textwidth]{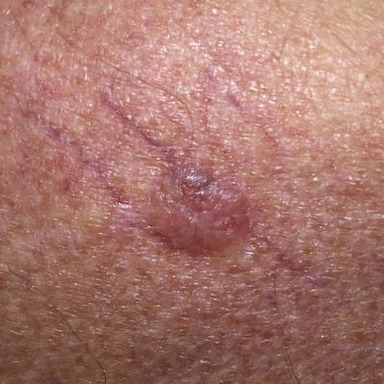}
                \caption{M. malignant}
            \end{subfigure}
            \vfill
            \begin{subfigure}[b]{0.45\textwidth}
                \includegraphics[width=\textwidth]{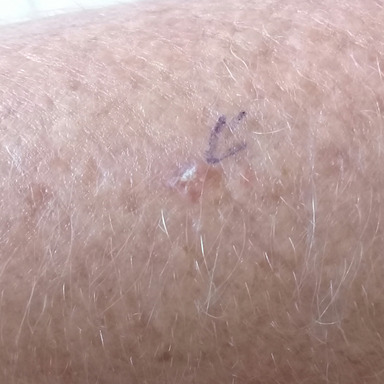}
                \caption{F. benign}
            \end{subfigure}
            \hfill
            \begin{subfigure}[b]{0.45\textwidth}
                \includegraphics[width=\textwidth]{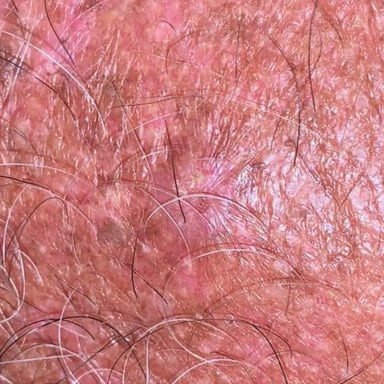}
                \caption{M. benign}
            \end{subfigure}
        \end{minipage}
        \caption{\rebuttal{Comparison of skin lesion images: The left panel shows ISIC dermoscopic images, the middle panel presents four representative lesions DERM7PT dermoscopic images, and the right panel displays PAD‑UFES‑20 smartphone‑captured images. In each panel, the top row contains malignant lesions from a male (M.) and a female (F.) patient, while the bottom row shows benign lesions from male and female patients, illustrating the visual characteristics across the different sources.}}
        \label{fig:skin_lesions_examples}
    \end{figure*}

    \rebuttal{We used three skin-lesion datasets: a curated ISIC subset (for training, validation, and internal testing), plus PAD-UFES-20 and DERM7PT (for external testing only). The ISIC subset was derived from the full archive after preprocessing (Section \ref{preprocessing_section}), with controlled demographic distributions for sex and age. Figure \ref{fig:skin_lesions_examples} illustrates representative samples. Dermoscopic images (ISIC and DERM7PT, respectively, in the left and middle panels) show greater detail and subsurface structures, particularly with polarised dermoscopy, potentially improving diagnostic accuracy (\cite{Kittler2002-kl}). Smartphone images (PAD-UFES-20) exhibit greater variation in lighting, angle, and background.}

    \subsubsection{Collection and preprocessing}\label{preprocessing_section}
    \paragraph{ISIC based dataset}
        We used the ISIC archive's gallery browser (\cite{Gutman2016-lf,Codella2017-rd,Codella2019-cn,Tschandl2018-sz,Combalia2019-jj,Veronica2021patient,Isic-query}), which contained 81,155 dermoscopic images of skin lesions with associated age and sex metadata. The archive was queried for dermoscopic images with diagnoses of "benign" or "malignant" in all age groups and both sexes, yielding 71,035 images (62,439 benign, 8,596 malignant). After data collection, we performed several preprocessing steps to ensure data quality. First, we removed cases lacking age attribute values, leaving 70,843 lesions (62,291 benign and 8,552 malignant). We then removed duplicate images by comparing MD5 hash-values, following Cassidy’s method (\cite{Cassidy2022-wx}). After duplicate elimination, 69,982 lesions remained (61,472 benign and 8,510 malignant). Finally, we identified multiple images of the same patient (multiplets) using patient ID attributes and excluded them, resulting in 35,884 lesions (28,810 benign and 7,074 malignant). \rebuttal{This removal reduces bias by preventing a single patient from disproportionately influencing the model and eliminates the risk of data leakage across train/validation/test splits.} Among the benign lesions, 13,207 were from female patients and 15,603 from male patients. Among malignant lesions, 3,012 were from female patients and 4,062 from male patients.
        
    \paragraph{PAD-UFES-20 based dataset}
        \begin{figure}[!ht]
            \centering
            \includegraphics[width=1\columnwidth] {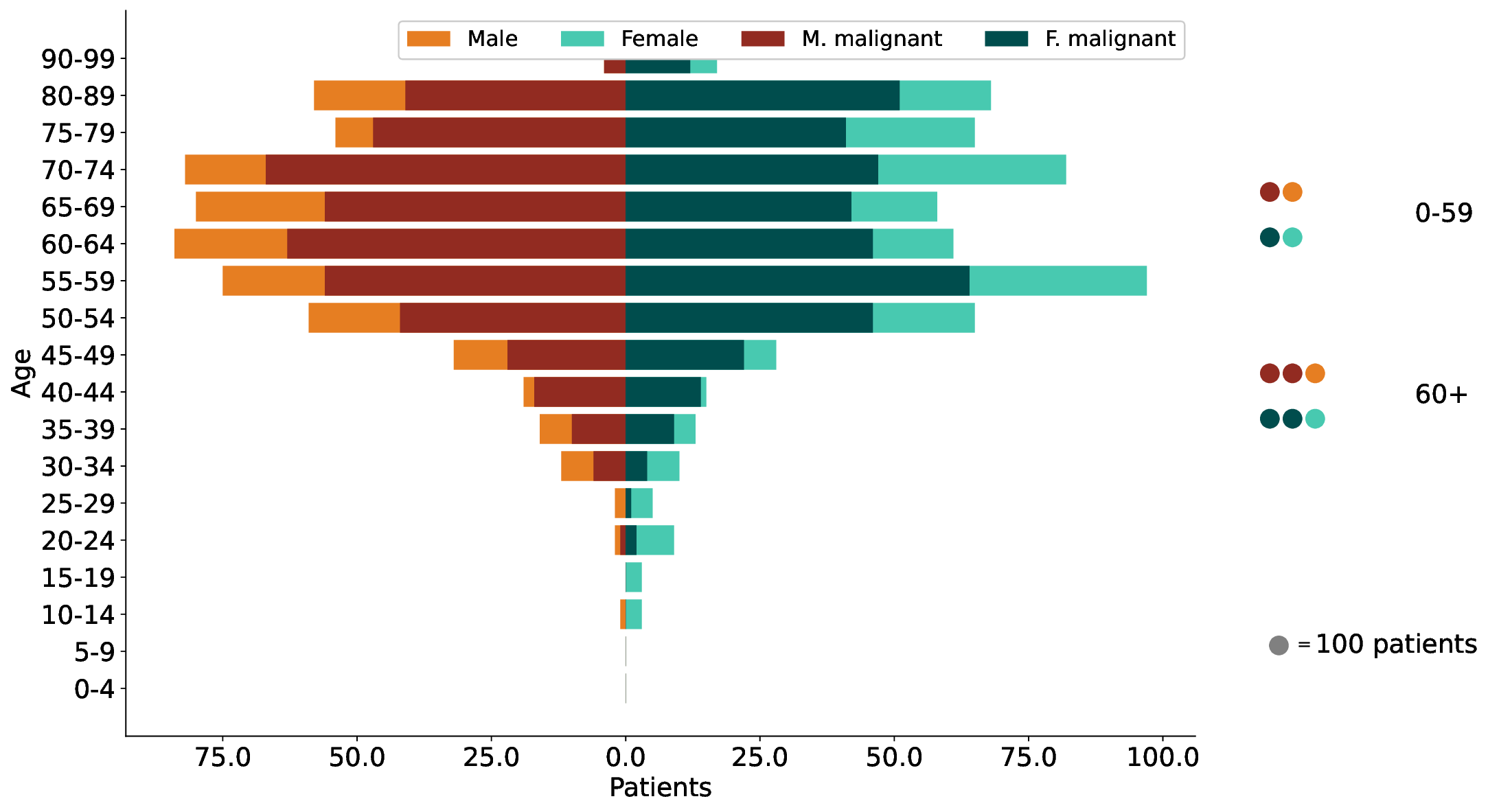}
            \caption{Age distribution across sex and diagnosis type in the curated \textbf{PAD-UFES-20} dataset, showing the breakdown between male and female patients with benign versus malignant skin lesions. For a detailed breakdown of lesion counts, see the Appendix \ref{pad_ufes_lesion_dist}.}
            \label{fig:age_dist_curated_padufes}
        \end{figure}
    
        For external validation, we used the PAD-UFES-20 dataset (\cite{pacheco2020pad}), which comprises clinical skin-lesion photographs taken with smartphones from patients in Brazil. The original collection comprised 2,298 records. To prepare the data for cross-validation, we performed a sequence of cleaning operations to ensure completeness, consistency, and nonredundancy. First, we removed all entries lacking a sex label (804 rows in total), leaving a fully sex-annotated cohort (741 male and 753 female patients). No records were missing age information, eliminating the need for further imputation. Next, we excluded any lesion without an associated biopsy result, thereby ensuring that every sample used for model evaluation had a definitive pathological ground truth; this filtering reduced the set to 1,179 cases. We then consolidated diagnostic labels into two broad categories: malignant (Melanoma, Basal Cell Carcinoma, Squamous Cell Carcinoma) and benign (Actinic keratosis, Nevus, and Seborrheic keratosis). Finally, we checked for duplicate entries representing the same patient-lesion pair and found none. The final curated dataset comprises a malignant subset of 432 male and 401 female patients, and a benign subset of 148 male and 198 female patients, resulting in a dataset of 1,179 unique records. Figure \ref{fig:age_dist_curated_padufes} illustrates the age distribution stratified by sex for both malignant and benign cases in the curated dataset.

    \paragraph{\rebuttal{DERM7PT based dataset}}
        \rebuttal{We used the publicly released dermoscopic collection (\cite{Kawahara2018-tx}), comprising 1,011 cases originally curated for the Interactive Atlas of Dermoscopy by Argenziano et al. (\cite{argenziano2000interactive}). Each case includes a dermoscopic image, a clinical image, patient metadata, and eight labels (seven 7-point checklist criteria plus diagnosis). Sex metadata is available for all samples, though age information is absent. We grouped diagnostic codes into two categories: benign lesions (nevi, dermatofibromas, lentigines, melanoses, vascular lesions, and seborrheic keratoses) and malignant lesions (basal cell carcinoma and all melanoma subtypes, including in situ, invasive, and metastatic). The malignant subset comprises 160 female and 134 male patients, while the benign subset includes 362 female and 355 male patients. For analysis, we utilised only the dermoscopic images.}

    \rebuttal{Table \ref{tab:overview_skin_lesion_datasets} provides an overview of the curated datasets used in this study, including their modalities, sample sizes, geographic origins, and demographic distributions.}

\subsubsection{Dataset creation}
    \rebuttal{We developed a method (\cite{Raumanns2025-nt}) to create diverse dataset compositions using linear programming (LP), a standard mathematical optimisation technique. Our pipeline consists of two steps: (1) generating a demographically controlled subset using an LP model, and (2) splitting it into training, validation, and hold-out test sets. We applied this exact pipeline to every experiment, whether adjusting the male-to-female patient ratio or reshaping the age-group distribution. We chose LP over random sampling. LP exactly satisfies multiple demographic constraints (sex, age, lesion diagnosis) while selecting the largest possible subset meeting those ratios. Random sampling approximates the target distribution but often discards or duplicates rare cases. It cannot guarantee specific subgroup constraints (e.g., dark-skinned males aged 50–60 with malignant lesions). The accurate and reproducible control provided by LP is therefore essential for rigorous bias and fairness analysis. It can also be easily extended with additional constraints that random sampling cannot accommodate. In what follows, we describe the specific steps used to create the datasets for each age- and sex-related experiment.}
    \paragraph{\rebuttal{Dataset composition using linear programming}}
    The goal of the LP model is to maximise the number of instances of skin lesions within defined constraints, as we express below:

\begin{equation*}
\begin{aligned}
& \text{Find a vector } x \text{ (decision variables)} \\
& \text{that maximises } f = x_1 \text{ (objective function)} \\
& \text{subject to } a_{i1}x_1 + a_{i2}x_2 + \dots + a_{in}x_n\leq b_i \text{ (constraints)} \\
& \hspace{2.5em} \text{for } i = 1, \dots, N_i \\
& \text{and } x_j \geq 0 \text{ (non-negativity constraints)} \\
& \hspace{2.5em} \text{for } j = 1, \dots, N_j
\end{aligned}
\end{equation*}

\rebuttal{The model has $N_j$ decision variables ($x_1,\dots,x_{N_j}$) and $N_i$ constraints. Each decision variable corresponds to specific categories (e.g., benign lesions in female patients aged $> 60$ years). The objective function maximises the count of malignant instances $x_1$. In the ISIC archive, there are fewer malignant instances than benign ones, and the goal is to achieve a balance between the two. The constraints enforce bounds on individual groups and maintain inter-group ratios. Representative groups include all benign lesions, all females over 60 years old, and all males under 60 years of age. A key constraint maintains class balance by ensuring an equal number of malignant and benign lesions ($x_1 - x_2 = 0$). Non-negativity constraints prohibit negative values for all decision variables. The complete LP formulation is detailed in Appendix \ref{lp_model_sex_dist}.}

\begin{table*}[hbt!]
\caption{\textbf{ISIC-based} datasets are distributed amongst malignant, benign, male patients (M), and female patients (F) categories for both training and validation. Bold value indicates the minimal malignant‑lesion count.}
\resizebox{\textwidth}{!}{  
\setlength{\tabcolsep}{3pt}
\begin{tabular}{llllllll}
& M100 & \rebuttal{F5M95} & F25M75 & F50M50 & F75M25 & \rebuttal{F95M5} & F100 \\
\hline
Malignant (M/F) & \textbf{2206} (2206/0) & 2322 (2206/116) &  2941 (2206/735) & 4412 (2206/2206) & 3235 (809/2426) & 2554 (128/2426) & 2426 (0/2426)\\
Benign (M/F) & 2206 (2206/0) & 2322 (2206/116) & 2941 (2206/735) & 4412 (2206/2206) & 3235 (809/2426) & 2554 (128/2426) & 2426 (0/2426)\\
\end{tabular}
}
\label{tab:result_lp_model}
\end{table*}

Within set constraints, the optimal solution maximises malignant lesions and assigns value to decision variables. To find this solution, we created a unique LP model for each dataset. Table \ref{tab:result_lp_model} shows the result of the LP model for the different datasets. We adopted a procedure to obtain the final solution for each distribution, consisting of the following steps. First, we solved the LP model to identify the optimal composition of a balanced test set while maximising the number of malignant lesions. From this balanced set, we reserved one‑eighth as a hold‑out test set. Second, we recalibrate the upper‑bound constraints using the lesion counts observed in the hold‑out set. With these updated bounds, we resolved the LP model to derive the final solutions for the various distributions. Third, after obtaining solutions from the LP model, we determined the minimum number of malignant instances in all datasets. Fourth, we scaled each dataset proportionally to the minimum value, preserving demographic distributions while ensuring comparability.\\

    \paragraph{Sex distribution analysis}
        We created \rebuttal{seven} distinct training and validation datasets to analyse sex-related biases with varying patient ratios of female (F) to male (M). Each of the \rebuttal{seven} dataset instances was created using a distinct random seed. For each seed, we first created a hold‑out test set; the remaining data were then shuffled and split strictly into an 80\% training subset and a 20\% validation subset while preserving the original demographic ratios. There was no overlap of lesions between the training and validation sets for any given seed, preventing overlapping samples from inflating performance. We maintained the same number of malignant and benign lesions in all datasets and balanced age distributions with the same numbers of patients below and above 60 years (median age) for each sex. 
        
        \rebuttal{The datasets consisted of a M100 set (100\% male patients), a F100 set (100\% female patients), a F95M5 set (95\% female, 5\% male patients), a F75M25 set (75\% female,25\% male patients), a F50M50 set (50\% female, 50\% male patients), a F25M75 set (25\% female, 75\% male patients), a F5M95 set (5\% female, 95\% male patients), and a separate balanced test set that matches the distribution of the F50M50 (equally‑split) dataset.}

\begin{figure*}[htbp]
    \centering
    \begin{subfigure}[b]{0.45\textwidth}
        \includegraphics[width=\textwidth]{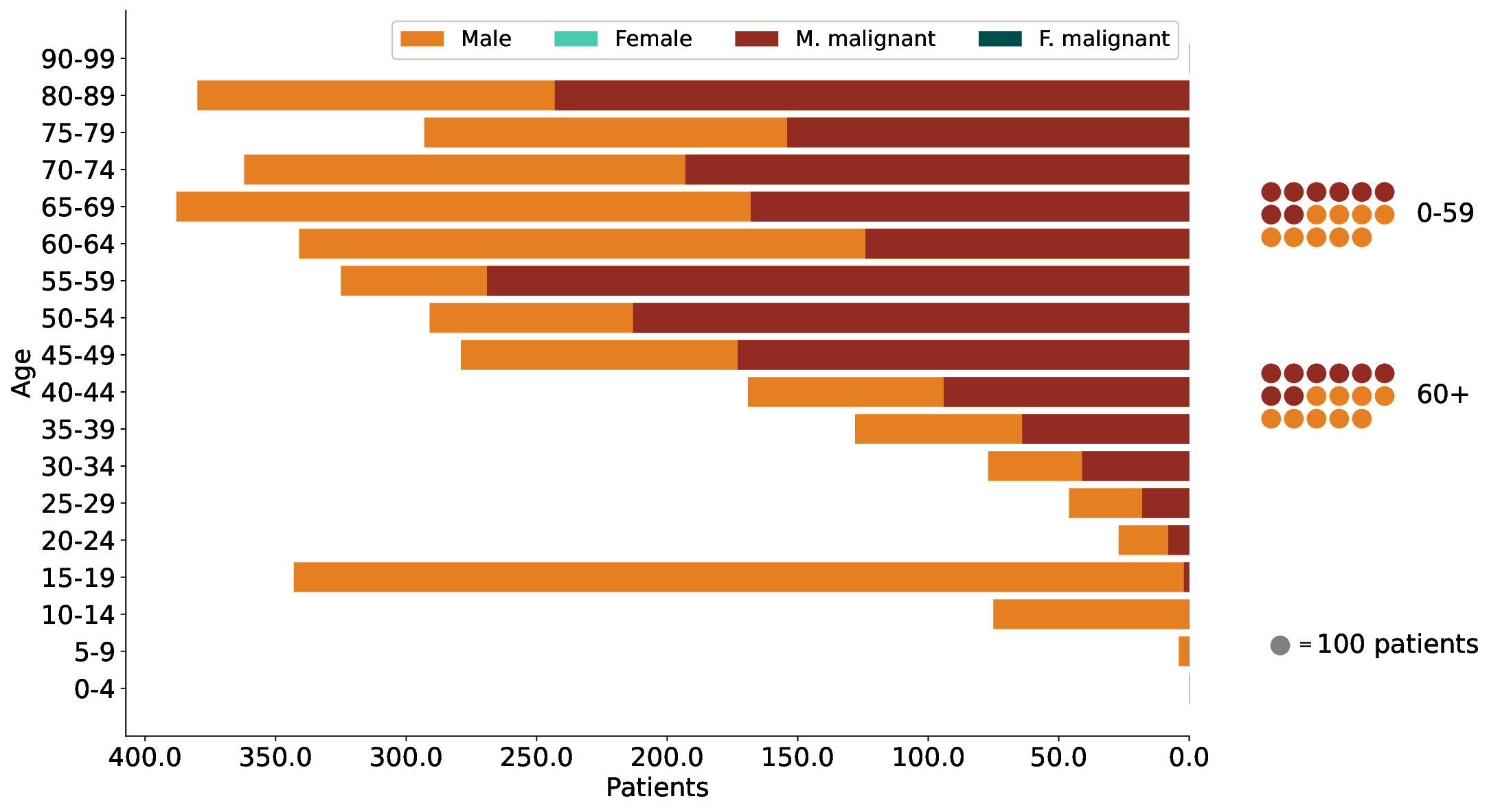}
        \caption{\MDD}
    \end{subfigure}
    \hfill
    \begin{subfigure}[b]{0.45\textwidth}
        \includegraphics[width=\textwidth]{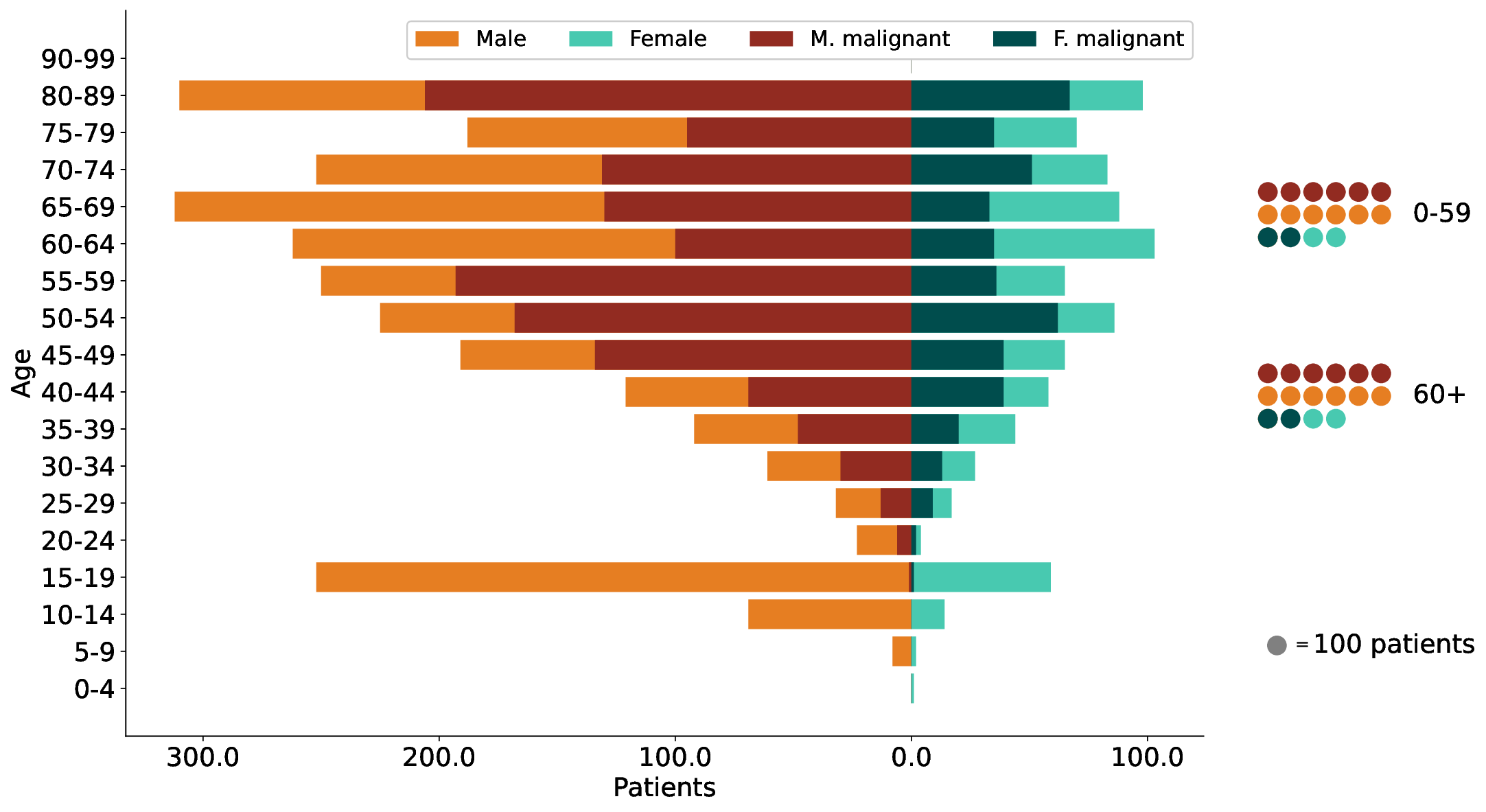}
        \caption{\MBD}
    \end{subfigure}
    
    \vspace{1em}
    
    \begin{subfigure}[b]{0.45\textwidth}
        \includegraphics[width=\textwidth]{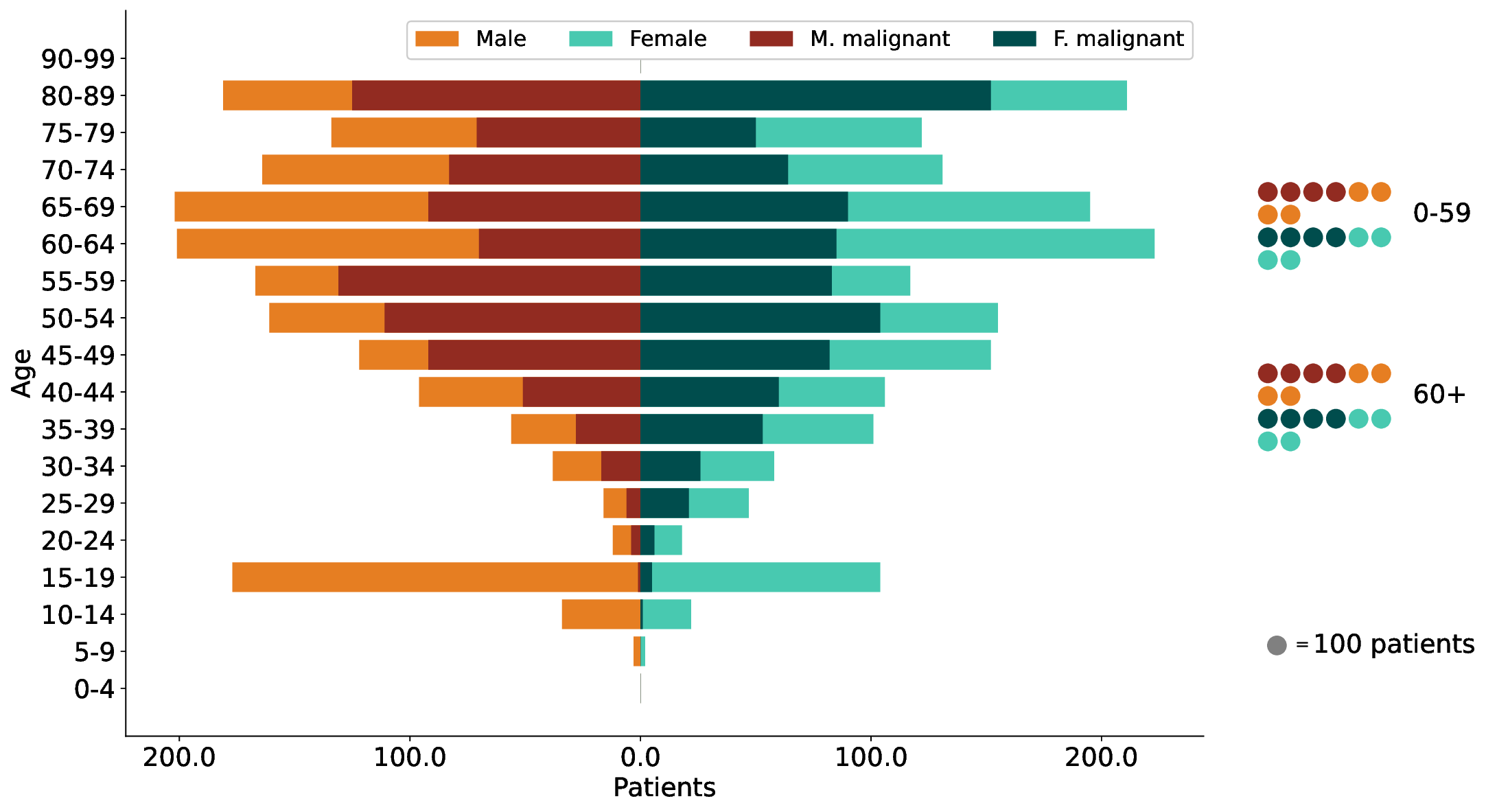}
        \caption{\ESD}
    \end{subfigure}
    \hfill
    \begin{subfigure}[b]{0.45\textwidth}
        \includegraphics[width=\textwidth]{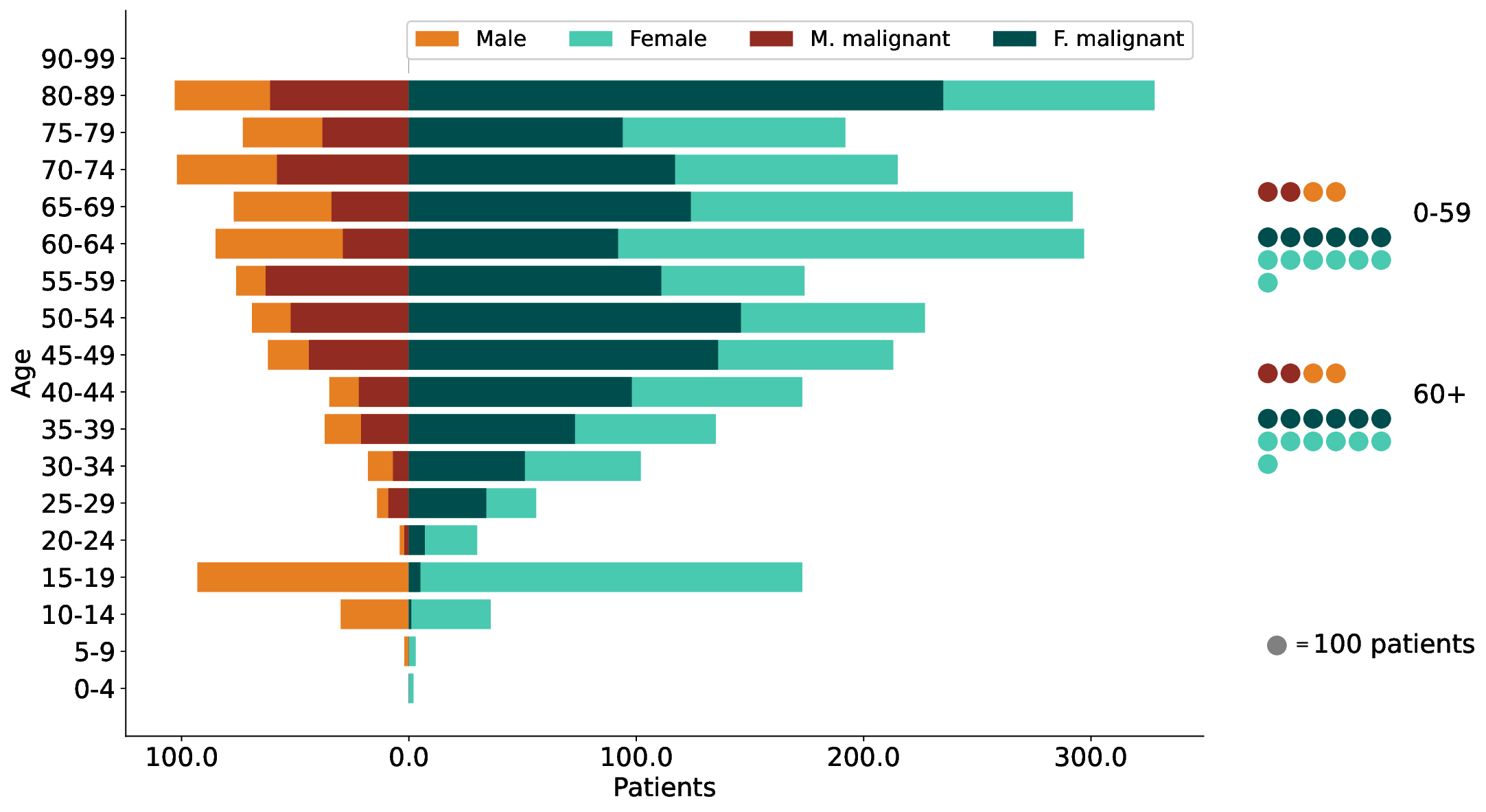}
        \caption{\FBD}
    \end{subfigure}
    
    \vspace{1em}
    
    \begin{subfigure}[b]{0.45\textwidth}
        \includegraphics[width=\textwidth]{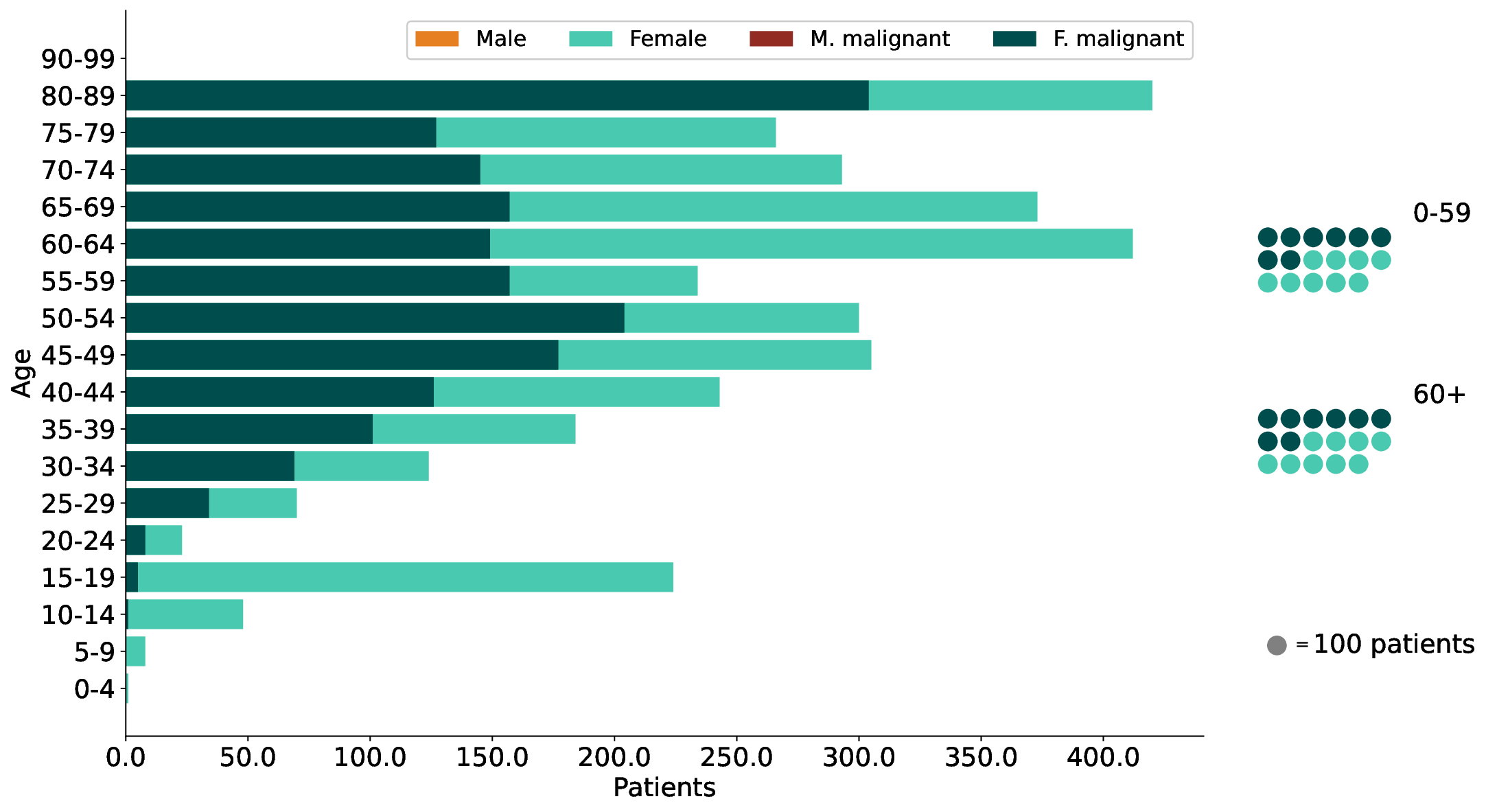}
        \caption{\FDD}
    \end{subfigure}
    \hfill
    \begin{subfigure}[b]{0.45\textwidth}
        \includegraphics[width=\textwidth]{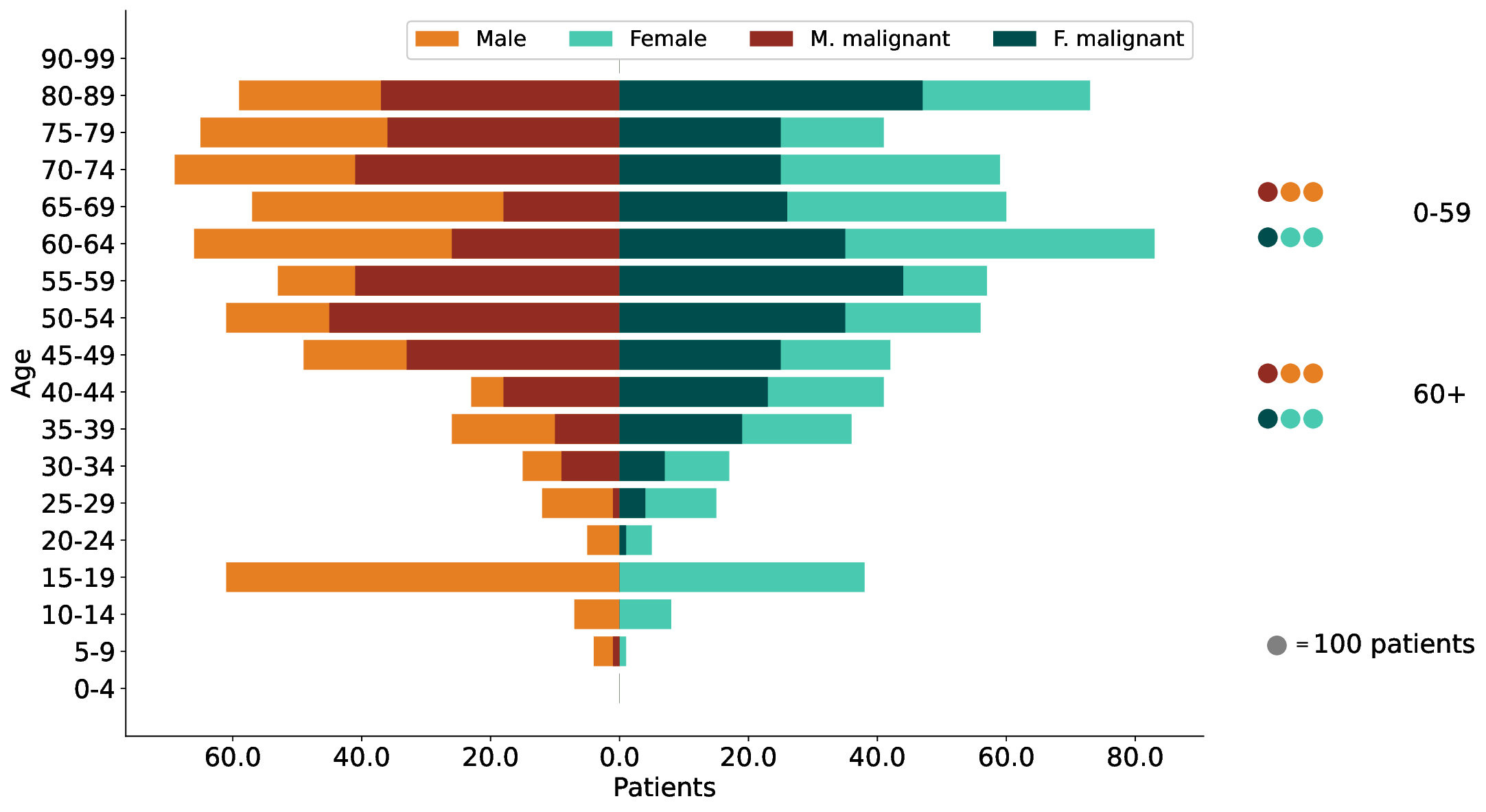}
        \caption{Balanced testset}
    \end{subfigure}
    \caption{Age distributions in the ISIC-based datasets range from M100 to F100 (a to e), with a balanced test set (f). The training sets (3,528 records, approximately 80\% of the total) and corresponding validation sets (880 records, approximately 20\%) span distributions from M100 to F100 and maintain similar population compositions. The total of 4,412 records ($2 \times 2,206$ lesions) reflects the inclusion of both malignant and benign cases, with the base value of 2,206 taken from Table \ref{tab:result_lp_model}. Test sets contain 1,264 records. The visualised distributions correspond to seed value 1970; distributions for other seeds are equivalent.}
    \label{fig:population_pyramids}
\end{figure*}

        \rebuttal{Figure \ref{fig:population_pyramids} illustrates the age distributions across the sex-based training datasets and the balanced test set.}

    \paragraph{Age distribution analysis}
        Similar to the sex‑distribution analysis, we built the training, validation, and test sets for the age‑focused experiments using an LP model (see Appendix \ref{lp_model_age_dist} for full details). We defined three age‑distribution schemes, each spanning the same five age brackets. Table \ref{tab:age_categories_distrib} shows the definition and proportions of the five age brackets ($A_{1}$–$A_{5}$) for each of the three schemes. In the YOUNGER scheme, the majority of samples come from the youngest age brackets, with the proportion gradually decreasing toward older age groups. The BALANCED scheme allocates samples uniformly across all five brackets. Finally, the OLDER scheme is the inverse of the younger‑skewed arrangement, concentrating most samples in the oldest age groups. Each distribution enforces a strict 1:1 balance between malignant and benign lesions and a 1:1 balance between male and female patients. For each scheme, we generated five independent instances using five different seeds. For each seed, we divided the data into non-overlapping training and validation sets and reserved a holdout test set with a uniform age category distribution (based on the BALANCED scheme). See Table \ref{tab:train_val_test_skin_lesions} for the count of skin lesions in each split.
        \begin{table}[hbt!]
\centering
\caption{Proportion of the five age groups ($A_{1}$–$A_{5}$) across the three schemes (YOUNGER, BALANCED, and OLDER) for the \textbf{ISIC-based} data. The age brackets are defined as follows, \rebuttal{where $a$ represents the patient's age in years: $A_1 = {0 \leq a \leq 50}$, $A_2 = {51 \leq a \leq 60}$, $A_3 = {61 \leq a \leq 70}$, $A_4 = {71 \leq a \leq 80}$, and $A_5 = {a \geq 81}$.}}
\small
\begin{tabular}{lccccc}
& $A_{1}$ & $A_{2}$ & $A_{3}$ & $A_{4}$ & $A_{5}$ \\
\hline
YOUNGER & 0.35 & 0.30 & 0.20 & 0.10 & 0.05  \\
BALANCED & 0.20 & 0.20 & 0.20 & 0.20 & 0.20 \\
OLDER & 0.05 & 0.10 & 0.20 & 0.30 & 0.35 \\
\end{tabular}
\label{tab:age_categories_distrib}
\end{table}        
        \begin{table}[hbt!]
\centering
\caption{Division of the in \textbf{ISIC-based} dataset into training and validation subsets for all three age‑distribution schemes. The numbers shown in the diagram indicate the count of skin‑lesion images in each split.}
\small
\begin{tabular}{lccc}
& Training & Validation & Testing \\
\hline
YOUNGER & 4740 & 1192 &   \\
BALANCED & 4120 & 1040 & 1020 \\
OLDER  & 2348 & 600\\
\end{tabular}
\label{tab:train_val_test_skin_lesions}
\end{table}

\subsection{Model}
    Using our carefully constructed datasets, we implemented three different architectures based on the ResNet50 model \rebuttal{(}\cite{He2016-ow}\rebuttal{)}. We selected ResNet50 for its proven performance in medical imaging and widespread adoption (\cite{Xu2023-cg}), enabling meaningful study comparisons. These architectures evaluate different approaches to handling demographic information:

\paragraph{The single-task baseline model}
    The single-task baseline model, enhanced with two fully connected layers, uses a sigmoid activation function and binary cross-entropy loss.

\paragraph{The multi-task reinforcing model} The multi-task ``reinforcing'' model with three layers added to the convolutional base, produces two outputs: one for classification and another for the demographic attribute (either sex or age, depending on the specific experiment). Please note that we use the term reinforcing here in the sense of “strengthening influence”, not in the reinforcement learning (RL) sense. When the attribute is sex (binary: male/female), we used a binary cross-entropy loss ($L_{c}$) and a sigmoid activation function for both heads. For age, which is treated as a continuous variable, we replace the binary loss with a mean-squared error loss applied to the normalised age value. Both heads (primary and demographic) receive equal weighting in the overall objective, while the classification head continues to use its standard loss function.
    When we use the auxiliary head for age prediction, we first normalise the age labels to the unit interval ([0, 1]) using the minimum and maximum ages observed in the dataset. During inference, we denormalise the sigmoid output back to the original age scale. Because the admissible age range is fixed and known a priori, this normalisation-denormalisation procedure preserves the semantic meaning of the prediction while keeping an identical model architecture for both auxiliary tasks.

\paragraph{The multi-task adversarial model}
    The multi-task adversarial model was implemented following the methodology of Adeli et al. and Abbasi-Sureshjani et al. (\cite{Adeli2021-da,abbasi2020risk}), using a network with a shared feature encoder and two classifier heads. One classifier targeted skin cancer classification; the other predicted confounders such as sex or age. We used the ResNet architecture to compare performance with baseline and reinforcement models in a systematic way.
    We trained the skin‑cancer classifier and its encoder with a standard cross‑entropy loss ($L_c$). The choice of bias‑predictor head loss ($L_{\text{bp}}$) depended on the demographic variable under study: for age‑distribution experiments we employed an age bias predictor head whose loss was defined as the negative‑squared Pearson correlation coefficient loss \rebuttal{(this worked for protected attributes that are continuous or ordinal (\cite{Adeli2021-da})}, while for sex‑distribution experiments we used $L_{\text{bp}}$ as a binary cross‑entropy loss, reflecting the binary nature of the attribute. 
    
    To reduce the predictiveness of the encoded features, we adversarially adjusted the encoder using a third loss term ($L_{\text{br}}$), with $\lambda$ governing the penalty for accurate demographic predictions: $L_{br}= \lambda{L_{\text{bp}}}$, following the practice of Abbasi-Sureshjani and colleagues (\cite{abbasi2020risk}).

\paragraph{Model training parameters}
    We optimised the single-task model using a grid search over random seeds, learning rates, and momentum values, selecting the combination that yielded the highest validation performance across all experiments.
    Subsequently, these optimised parameters were then applied to the reinforcement and adversarial models for a fair comparison. The optimal hyperparameters, selected via validation set performance, are as follows:
        \begin{equation*}
        \begin{aligned}
        & \text{Pre-training:} & \text{ImageNet} \\
        & \text{Input size:} & 384 \times 384 \text{ pixels} \\
        & \text{Max epochs:} & 40 \\
        & \text{Batch size:} & 20 \\
        & \text{Learning rate:} & 2.0 \times 10^{-5}
        \end{aligned}
        \end{equation*}

   To mitigate overfitting and improve model robustness, we implemented an early stopping technique (patience = 10 epochs) and data augmentation techniques. Following prior implementations in the literature, we implemented our baseline and reinforcement models in Keras with the TensorFlow backend (\cite{Geron2022-bm}), while our adversarial model was implemented in PyTorch (\cite{Paszke1912-ny}) to maintain consistency with existing adversarial learning frameworks.

\section{Experiments}\label{Experiments}
    \subsection{Demographic bias evaluation}
\begin{figure*}[!ht]
                \centering
                \includegraphics[width=0.7\textwidth] {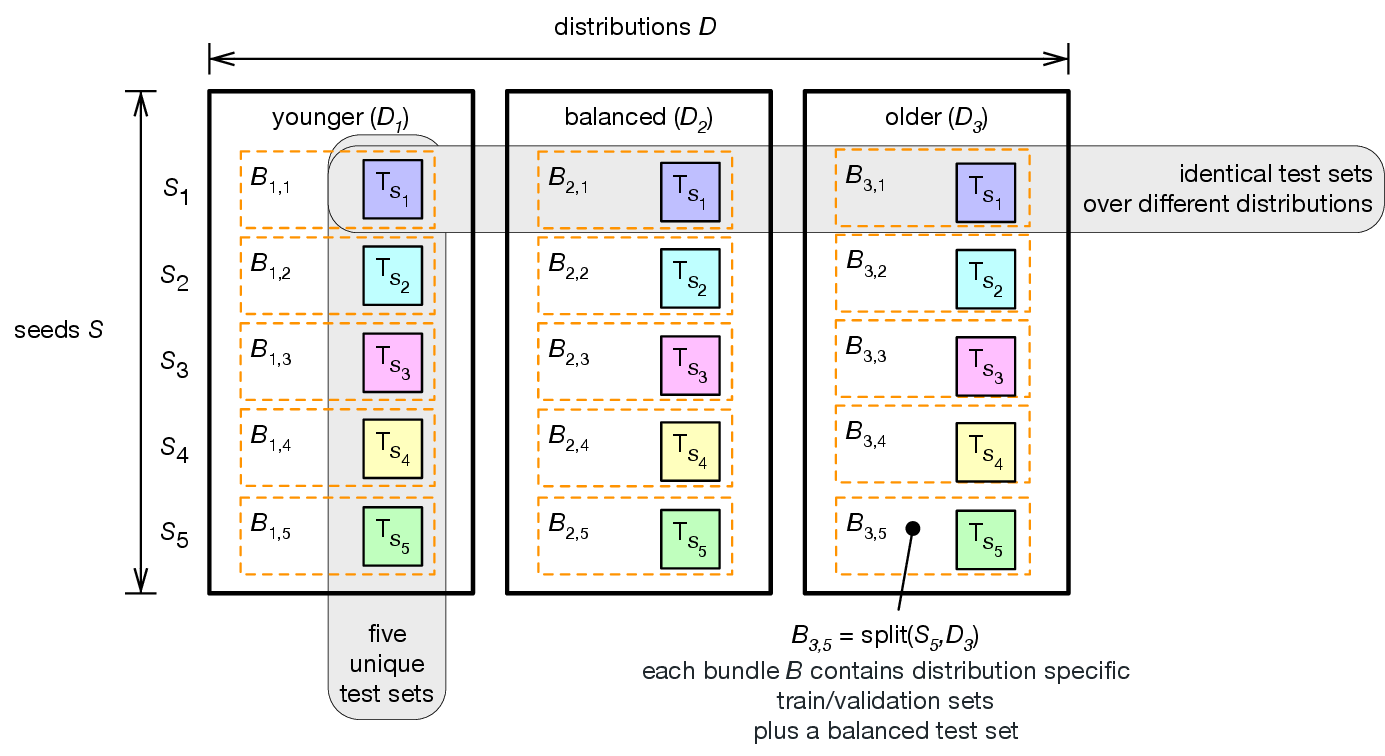}
                \caption{\rebuttal{Experimental workflow (age analysis; sex analysis is analogous): for each distribution $D$, we build five independent bundles $B$, each split into mutually exclusive test, training, and validation sets with a fixed seed $S$. The same test set is shared by all distributions for a given seed, and training/validation splits use the same seed. The distribution properties of the training/validation splits are consistent with the distribution under test. All three models are evaluated on this test set, allowing within‑ and cross‑distribution comparisons, and the process is repeated with multiple seeds to reduce random effects.}}
                \label{fig:seed_splits}
            \end{figure*}

    \paragraph{\rebuttal{Defining test, training and validation bundles}}
            \rebuttal{We took steps to ensure that any observed performance differences between models are attributable to the models themselves and the underlying data distributions, rather than to inconsistencies in how we split the data. By fixing a single random seed for each run, we use the same hold‑out test set across distributions, making cross-distribution comparisons meaningful. Importantly, we deliberately construct the hold‑out test set to be balanced across relevant subgroups (such as age or sex). This balanced test set removes bias toward any particular segment and allows us to isolate the effect of bias in the training data itself. We acknowledge, however, that measuring “true” behaviour across the entire population would require a test set whose distribution matches the expected real‑world population; our balanced test set is chosen specifically to evaluate bias rather than to predict real‑life performance. When we generate the training and validation partitions with the same seed, we ensure that these splits faithfully reflect the target distribution. By repeating the entire process with several different seeds, we reduce the influence of random variation in the data. Implementation is illustrated in Figure \ref{fig:seed_splits}. Our experimental workflow enables both within‑distribution and cross‑distribution model comparisons.}

    \paragraph{Sex distribution} 
        \rebuttal{For a thorough evaluation, we created five bundles for each of the seven distributions: F100, F5M95, F25M75, F50M50, F25M75, F5M95, and M100, resulting in 35 bundles (seven distributions $\times$ five seeds). We assessed AUC overall and within male and female subgroups for each learning strategy and dataset combination. Using three learning strategies on 35 bundles, we conducted 105 experiments to evaluate model performance within and across all seven distributions. Notably, multi-task models cannot unlearn constant protected attributes (as in M100 and F100 experiments). These edge cases serve to stress-test the training pipeline's stability when demographic attributes are absent.}
    \paragraph{Age distribution}
        \rebuttal{In our age experiments, we conducted comprehensive evaluations using age-stratified data sets. We generate three distinct data distributions, across five age categories ($A_1-A_5$): YOUNGER (predominantly young patients), BALANCED (evenly distributed age categories), and OLDER (predominantly elderly patients). For each configuration, we evaluated the three model architectures. Performance was measured using AUC scores, with results visualised across different age distributions and model architectures. We conducted 45 experiments in total (15 bundles and three learning strategies) to evaluate model performance within and across the three distributions.}

\subsection{Reinforcing model: Auxiliary head analysis}
        \rebuttal{We evaluate the auxiliary prediction head of the reinforcing model to serve two purposes: (1) assessing whether the multi-task architecture successfully learns the demographic signal, and (2) providing a mechanistic explanation for bias mitigation in the primary task. Specifically, if the auxiliary head fails to learn the demographics, the reinforcing model loses its regularisation effect, which may explain observed failures in bias mitigation. An auxiliary-head analysis was omitted for the adversarial model because the original network implementation did not provide demographic outputs.}

        \paragraph{Evaluation of the sex‑prediction head}
            We evaluated the auxiliary sex‑prediction head of the multi‑task reinforcing model using two complementary metrics. First, we computed the AUC to assess discriminative ability. Second, we measured the Brier score (\cite{Rufibach2010-tb}) separately for male and female patients. We interpret the Brier score ($\beta$) for binary classification as follows: strong calibration for $0 \leq \beta < 0.05$, moderate for $0.05 \leq \beta < 0.15$, weak for $0.15 \leq \beta < 0.25$, poor for $0.25 \leq \beta < 0.35$, and very weak for $\beta \geq 0.35$. Lower Brier scores indicate tighter alignment between predicted probabilities and observed outcomes, whereas higher values reflect increasingly poor calibration. We do not evaluate the two edge cases (\FDD, \MDD) with a training set containing only one sex; the auxiliary head cannot learn a useful decision boundary.

        \paragraph{Evaluation of the age‑prediction head}
            For the auxiliary age‑prediction head of the multi-task reinforcing model, we used mean absolute error (MAE) as our primary performance measure. MAE computes the average absolute difference between the predicted age and the ground-truth age across all test samples. To disclose any systematic biases throughout the age spectrum, we compute MAE for the five age categories ($A_{1}$–$A_{5}$, see Table \ref{tab:age_categories_distrib}). Furthermore, to assess the quality of the age predictions, we computed the Pearson correlation coefficient ($\rho$) between the predicted ages and the ground-truth ages. This metric quantifies the linear relationship between the two variables and complements the MSE by indicating how well the model captures age trends. We interpret the Pearson correlation coefficient as follows: strong correlation for $0.5 \leq \rho < 1$, moderate correlation for $0.3 \leq \rho < 0.5$, and weak correlation for $0 \leq \rho < 0.3$. Only correlations with $p < 0.05$ were retained for further analyses.\\

\subsection{Cross-dataset evaluation}
      \rebuttal{To validate our findings and assess generalisability, we performed additional experiments using two external skin‑lesion datasets: PAD‑UFES‑20 (both sex and age) and DERM7PT (no age information). Using the saved weights from our previously trained models (base, reinforcement, and adversarial), we evaluated them on the external datasets without any further fine-tuning, using the same evaluation metrics. This cross-dataset validation approach provides insight into the robustness and transferability of our models across different patient populations and data collection contexts.}

\subsection{Exploratory performance assessment}\label{stat_analysis}
    We did not conduct formal statistical testing because our study is exploratory, and the number of observations is minimal. Applying p‑value–based tests to the model performances would yield unstable estimates that provide little trustworthy insight into whether any actual effect exists. Consequently, we refrained from labeling results as “significant” or “non‑significant.” As Amrhein et al. highlighted, such dichotomous labeling often leads to misinterpretation, and non‑significant findings are frequently mistaken for evidence of no effect (\cite{Amrhein2019-hl}). Instead, we present AUC values as boxplots and provide descriptive comparisons across subgroups, allowing readers to assess the magnitude and direction of any observed differences.

\section{Results}\label{Experimental results}

\subsection{Sex-specific model assessment}
\begin{figure*}[!ht]
    \centering
    \includegraphics[width=1\textwidth] {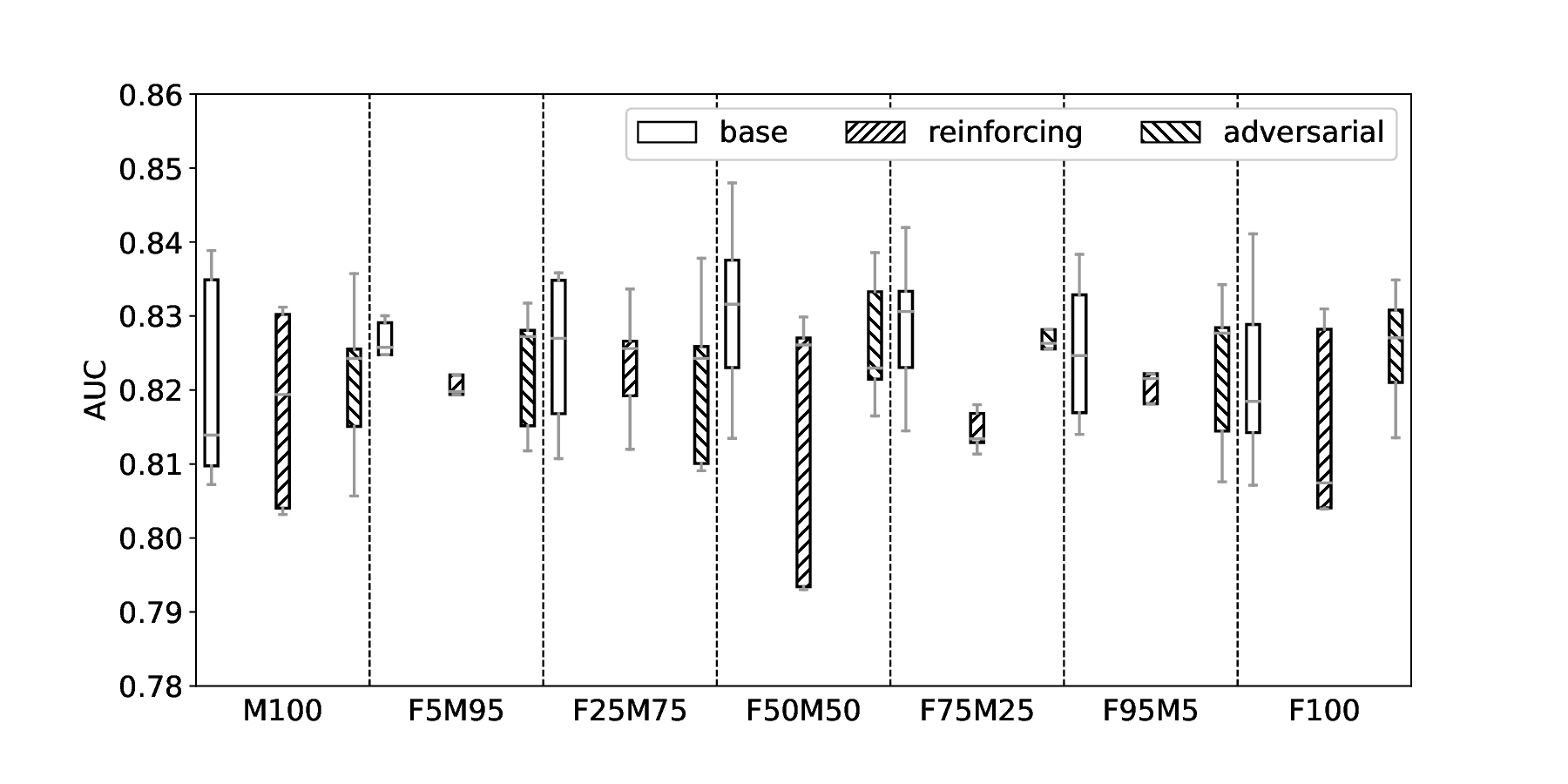}
    \caption{Comparison of model performance across sex distributions using datasets generated from the curated \textbf{ISIC dataset}. Box plots display AUC metrics for three model architectures (base, reinforcing, and adversarial) trained and validated on sex-biased datasets (M100, \rebuttal{F5M95},F25M75, F50M50, F75M25, \rebuttal{F95M5}, F100) and evaluated on a balanced dataset.}
    \label{fig:auc_dsitribution_sex}
\end{figure*}

\begin{figure*}[!ht]
    \centering
    \includegraphics[width=1\textwidth] {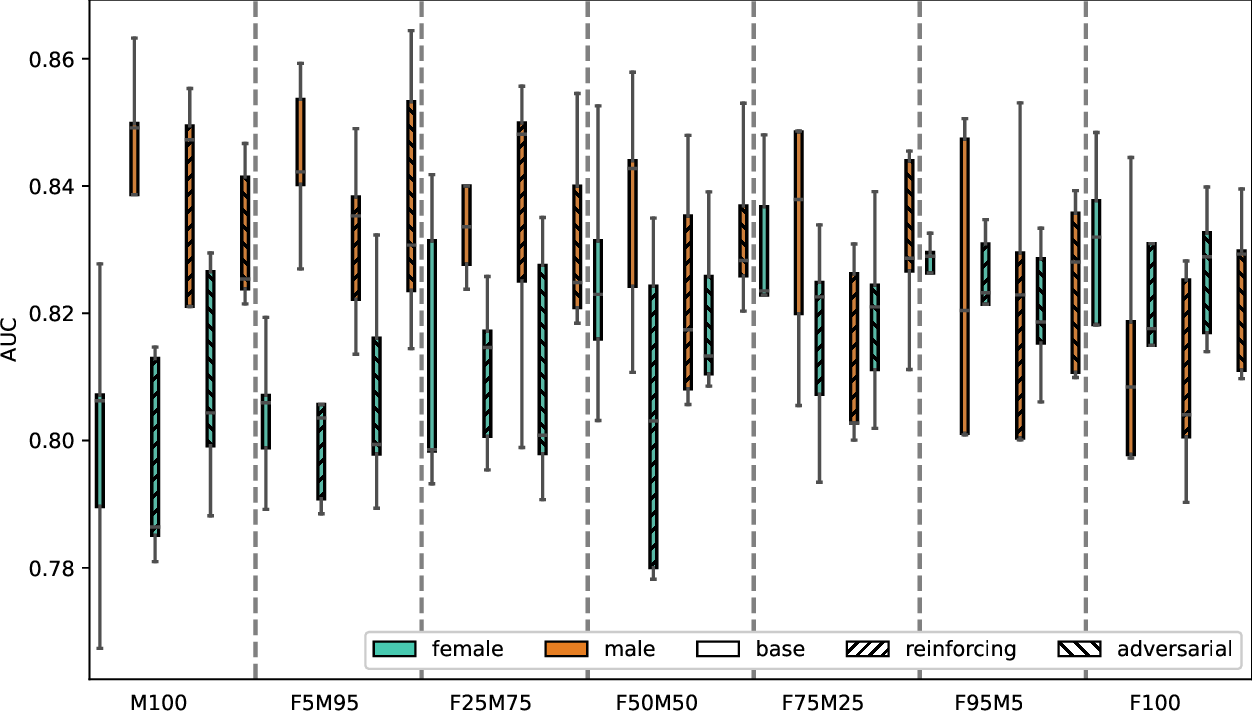}
    \caption{The AUC score varies based on data splits ranging from only male patients (\MDD) to only female patients (\FDD) in the \textbf{ISIC dataset}. We show base, reinforcing and adversarial model performance for female and male patient subgroups.}
    \label{fig:boxplot_reinforcing_adversarial_task}
\end{figure*}

    Figure \ref{fig:auc_dsitribution_sex} shows the performance of the model in sex distributions using box plots of AUC metrics for three types of models (base, reinforcement, adversarial). Figure \ref{fig:boxplot_reinforcing_adversarial_task} shows the impact of dataset distributions on three learning strategies, reporting AUC scores for both sexes.

    \paragraph{Comparable performance at model level}
        Our analysis in Figure \ref{fig:auc_dsitribution_sex} demonstrates that all three learning strategies achieve similar levels of effectiveness, showing only slight performance variations. Across the three model architectures, the accuracy scores maintain a consistent range between 0.79 and 0.85. Although there are minor variations between the approaches, \rebuttal{none clearly outperforms the others}.

    \paragraph{Sex-specific training data yields better results} 
        Figure \ref{fig:boxplot_reinforcing_adversarial_task} shows that all models perform better for male patients in male-only \rebuttal{ (M100), predominantly male (F5M95) and lightly male-skewed (F25M75)} scenarios. The reinforcing and base models show the most pronounced performance gap in the \rebuttal{predominantly} male dataset. The base and reinforcing show equal performance between subgroups in the balanced, lightly female-skewed \rebuttal{ (F75M25) and predominantly female (F95M5) scenarios}. The base model shows better performance for female patients in the female-only \rebuttal{(F100)} scenario, while the \rebuttal{reinforcing} and adversarial models show equal performance between male and female patients. Thus, our models appear more attuned to male patients in mixed-sex training sets, regardless of the percentage of female patients.
        The best results are achieved when both sexes are trained exclusively on their respective data. We hypothesise that a model trained on a single‑sex dataset may specialise in those sex-specific cues and often attains higher accuracy. In mixed‑sex training, the network must accommodate both distributions, which can lead to a modest performance dip, typically favouring the more dominant signal.

    \paragraph{Base model reveals sex bias} 
        We found substantial sex bias in the performance of the base model (see Figure \ref{fig:boxplot_reinforcing_adversarial_task}). In the male-only and \rebuttal{predominantly male} scenarios, we observed a substantial performance gap between male and female patients. \rebuttal{In the female-only scenario, there is also a performance gap; however, this is less pronounced.} \rebuttal{We found that the base model performed comparably for male and female patients across balanced, lightly skewed, and predominantly female experiments.} We assume the base model binds onto the most prevalent cues in the training set. When the data consist of only one sex or are heavily skewed, it reveals sex-specific visual patterns (e.g., hair density, skin texture) that aid classification, performing well for the majority sex but poorly for the minority. In a balanced, only mildly skewed female dataset, both sexes are equally represented, forcing the model to rely on lesion-intrinsic features for discrimination, leading to comparable accuracy for males and females.

    \paragraph{Reinforcement model partially successful in sex bias mitigation} 
        When we trained the model on \rebuttal{male majority} data, we observed performance disparities between the sexes (see Figure \ref{fig:boxplot_reinforcing_adversarial_task}). \rebuttal{With balanced training data, the reinforcement model successfully mitigates sex-based bias. Notably, this same bias reduction effect is observed in female-majority training sets as well.} 
        \rebuttal{We hypothesise that the reinforcing multi‑task model can reduce sex bias only when its auxiliary sex‑prediction head receives sufficiently informative female patient-related signals. In the only-male and predominantly-male scenarios, the encoder overfits to male-specific cues because the auxiliary head lacks enough female examples to learn a meaningful discriminator. Conversely, with balanced or mostly-female settings, the auxiliary head can learn a reliable sex classifier; its loss then regularises the shared encoder towards sex-invariant representations, thereby reducing bias.}

    \paragraph{Adversarial model reduces sex bias \rebuttal{in predominantly female training scenarios}.} 
        The adversarial model reduces sex bias in scenarios with predominantly female patients but is less effective in other scenarios, often favouring male patients. Its performance varies between experiments and datasets (see Figure \ref{fig:boxplot_reinforcing_adversarial_task}). \rebuttal{We suspect that complex anatomical confounders, such as variations in body hair distribution or skin texture, continue to act as strong proxies for sex in mixed populations, resisting the adversarial removal of these features.}

\begin{figure*}[!ht]
    \centering
    \includegraphics[width=1\textwidth] {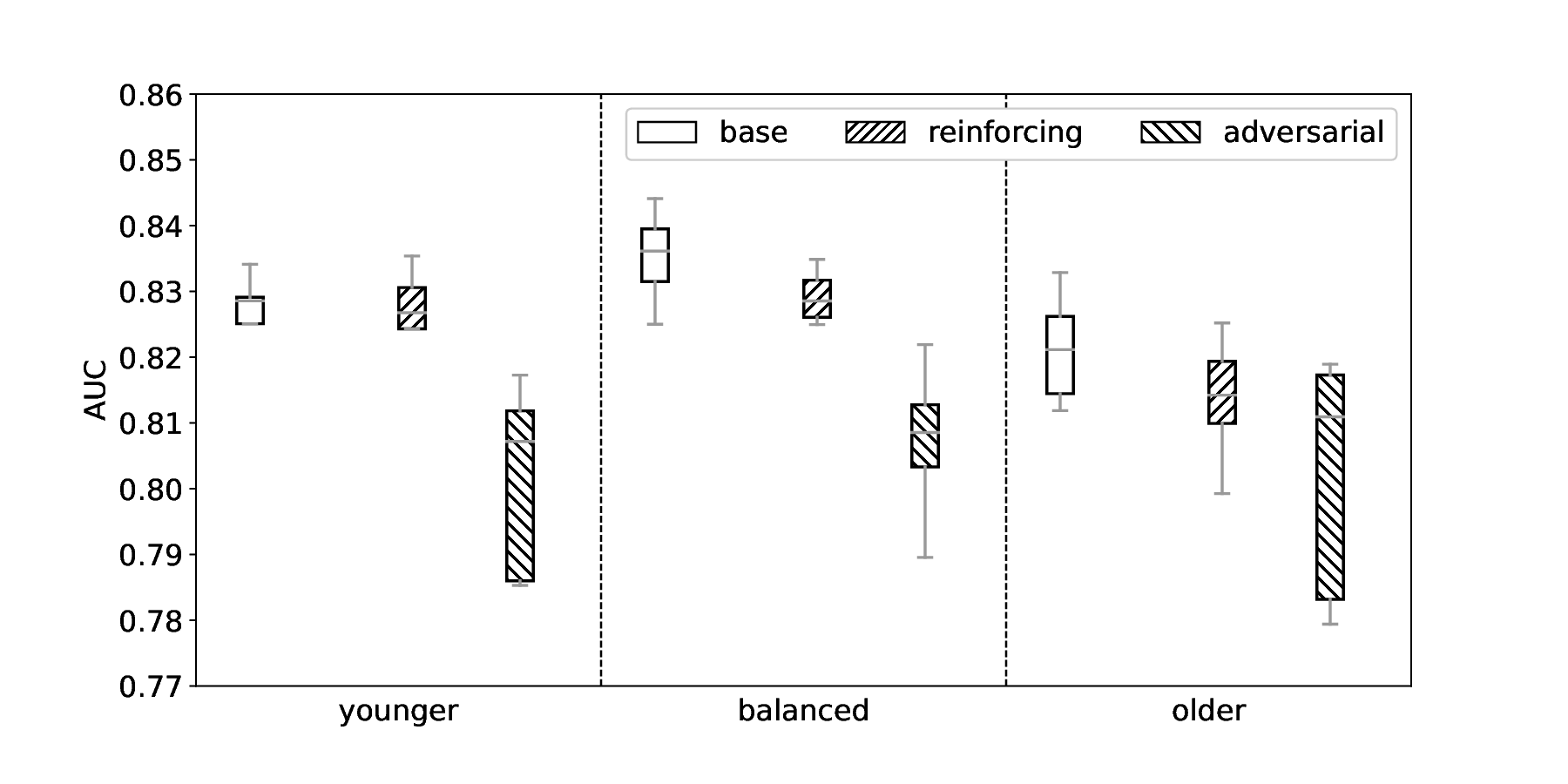}
    \caption{  Comparison of model performance across different age distributions using curated \textbf{ISIC dataset}. Box plots show the AUC metrics of three model architectures (base, reinforcing, and adversarial) trained and validated on age-biased datasets (YOUNGER, BALANCED, OLDER) and evaluated on a balanced test set.}
    \label{fig:auc_dsitribution_age}
\end{figure*} 

\begin{figure*}[!ht]
    \centering
    \includegraphics[width=1\textwidth] {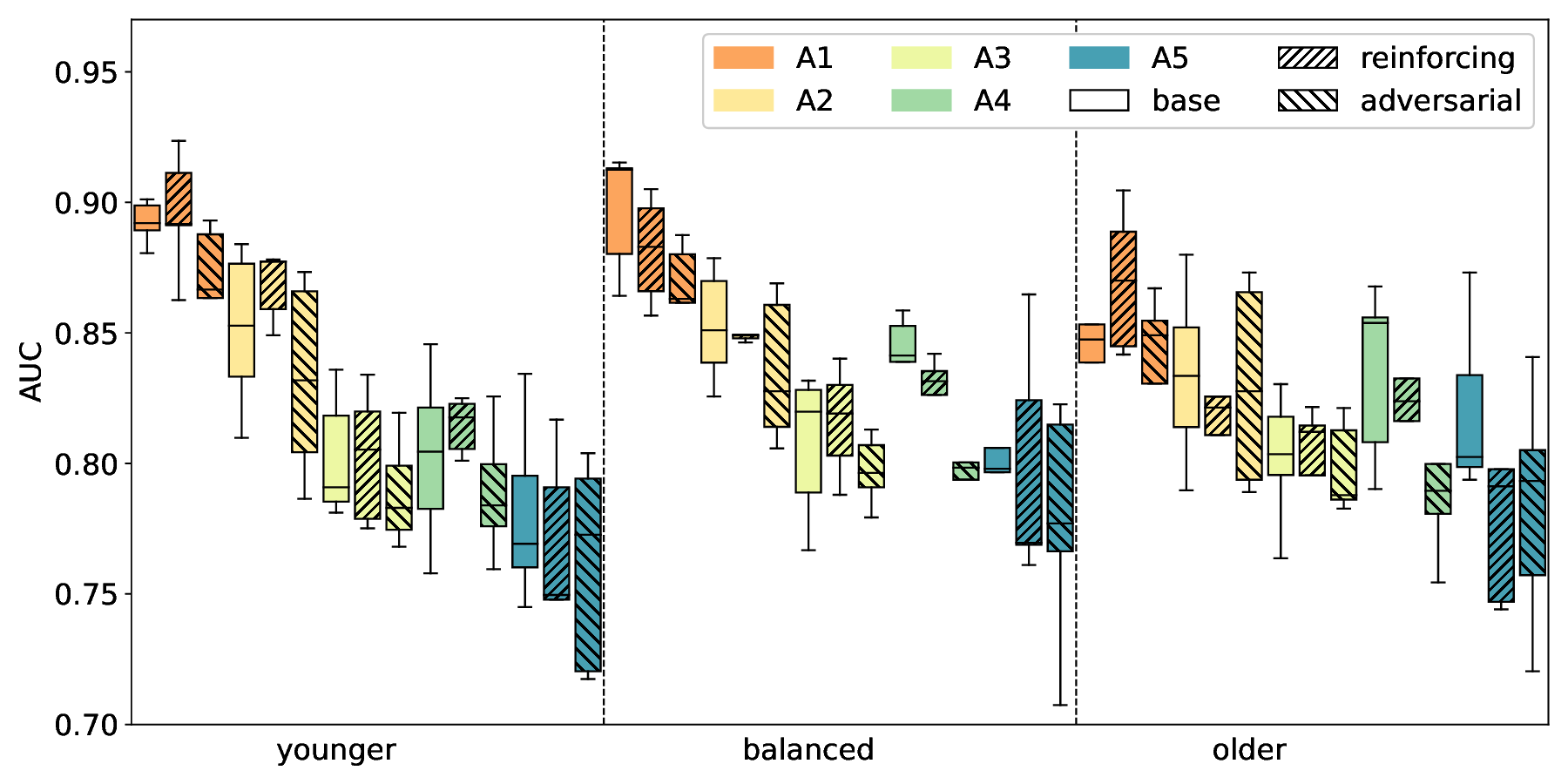}
    \caption{Age-stratified (using five age categories ($A_1-A_5$)) model evaluation across datasets with varying age distributions based on the curated \textbf{ISIC dataset}. The analysis compares AUC scores for three model architectures (base, reinforcing, and adversarial) using age-biased training sets (YOUNGER, BALANCED, OLDER). Each model was evaluated on a balanced test set, showing performance variations across different age distributions. \rebuttal{The age brackets are defined as follows, \rebuttal{where $a$ represents the patient's age in years: $A_1 = {0 \leq a \leq 50}$, $A_2 = {51 \leq a \leq 60}$, $A_3 = {61 \leq a \leq 70}$, $A_4 = {71 \leq a \leq 80}$, and $A_5 = {a \geq 81}$.}}}
    \label{fig:auc_subhroups_age}
\end{figure*}

\subsection{Age-specific model assessment}
    Figure \ref{fig:auc_dsitribution_age} presents a comparative analysis using box plots to illustrate AUC metrics across YOUNGER, BALANCED, and OLDER datasets. For a more granular understanding, Figure \ref{fig:auc_subhroups_age} provides an age-stratified evaluation using five distinct age categories ($A_{1}$–$A_{5}$), demonstrating how each model architecture performs when trained on differently distributed datasets and evaluated against a balanced test set.

    \paragraph{Comparable overall model performance}
        Looking at the overall performance of the model in all experiments (Figure \ref{fig:auc_dsitribution_age}), the adversarial model shows the highest variance in performance in different seeds compared to the other two strategies, particularly in the YOUNGER and OLDER cases. All three model approaches demonstrate comparable base performance levels with AUC scores falling within a 0.06 range, with both the base and reinforcement models showing a slight advantage compared with the adversarial model.

    \paragraph{Declining trend across categories}
        \rebuttal{As shown in Figure \ref{fig:auc_subhroups_age}, all three distributions show a decreasing AUC trend, with the youngest age bracket achieving the highest AUCs and the oldest the lowest. In the balanced distribution, the decreasing trend does not hold for the $A_4$ age bracket in either baseline or reinforcing models. $A_4$ performance is higher in the balanced case than in the younger-age distribution. In the older age distribution, $A_1$ bracket models show a slight performance decrease compared to the younger and balanced distributions. This decline is most pronounced for the baseline and adversarial models.} We assume the performance drop with increasing age is due to the nature of older skin, which exhibits more heterogeneous visual traits (such as wrinkles, pigment changes, and vascular alterations) that mask lesion cues.

    \paragraph{Strong performance for younger age categories}
        Models trained on the balanced dataset show the highest AUC for age category $A_{1}$, but experience a small performance drop for age category $A_{2}$ compared to models trained on the YOUNGER dataset (see Figure \ref{fig:auc_subhroups_age}). In contrast, the AUC values for the age categories $A_{3}$, $A_{4}$, and $A_{5}$ generally fall below 0.85. We hypothesise that the model excels in the youngest age group because young skin presents the most explicit lesion cues, with fewer wrinkles, pigment variations, or vascular artefacts to obscure the diagnostic signal, and the balanced training set supplies fewer examples of these clean patterns.

    \paragraph{Performance improvement of balanced  models} 
        Figure \ref{fig:auc_subhroups_age} illustrates that the base and reinforcing models trained on a balanced dataset show improved performance for the $A_{4}$ age category compared to the same models and age category trained in the younger dataset. We assume the balanced set provides enough examples that force the encoder to learn features that work across age groups. In the younger‑skewed data the model overfits to smooth, youthful textures, so its performance drops on the $A_4$ group. Adding balanced age samples thus improves AUC for that category.

\subsection{Sex-prediction head evaluation}
    \begin{table}[ht]
\centering
\caption{\rebuttal{Performance evaluation of the auxiliary sex-prediction head within the reinforcement model. Metrics include overall AUC and Brier scores ($\beta$), computed separately for male and female patient subgroups. Bold values denote optimal performance per column (highest AUC, lowest Brier score).}}
\small
\begin{tabular}{llll}
\hline
Training& AUC & $\beta$ &\\
dataset & & Female & Male\\
\hline
    \ESD & $\textbf{0.732} \pm \textbf{0.012}$ & $0.234 \pm 0.049$ & $0.238 \pm 0.033$ \\
    \rebuttal{F5M95} & $0.644 \pm 0.012$ & $0.959 \pm 0.013$ & $\textbf{0.001} \pm \textbf{0.000}$ \\
    \MBD & $0.682 \pm 0.043$ & $0.730 \pm 0.132$ & $0.028 \pm 0.021$ \\
    \FBD & $0.691 \pm 0.014$ & $0.040 \pm 0.012$ & $0.645 \pm 0.086$ \\
    \rebuttal{F95M5} & $0.612 \pm 0.012$ & $\textbf{0.001} \pm \textbf{0.001}$ & $0.948 \pm 0.026$ \\
\hline
\end{tabular}
\label{tab:sex_pred_analysis_reinforcing}
\end{table}

\rebuttal{Table \ref{tab:sex_pred_analysis_reinforcing} reports the auxiliary sex-prediction head's performance in the reinforcing model, including overall AUC and Brier scores for male and female subgroups. Predictive performance was limited across all skewed distributions but reached reasonable accuracy in the balanced case (AUC = 0.732), indicating that the encoder successfully learns sex-related features when the data is balanced. This aligns with the observed reduction in sex bias for the reinforcing model under balanced settings.
However, in skewed scenarios (e.g., F5M95), the auxiliary head's performance becomes highly asymmetric: it predicts the majority sex with high confidence (a weak Brier score) but fails on the minority sex (a strong Brier score). This inability to learn a robust sex discriminator explains why the reinforcing model cannot effectively regularise the encoder under skewed distributions, leading to persistent bias gaps (see F5M95 in Figure \ref{fig:boxplot_reinforcing_adversarial_task}).}

\paragraph{Highest AUC and \rebuttal{weak} Brier for the balanced training set}
    When we train the reinforcing model with equal numbers of male and female patients, the auxiliary sex‑prediction head learns a discriminative representation and achieves the highest AUC among the experiments. Brier scores for the \rebuttal{F50M50} case are \rebuttal{weak} for female and male patients (see Table \ref{tab:sex_pred_analysis_reinforcing}).

\paragraph{Strong Brier for the majority, very weak for the minority}
    When we train on a majority of male patients (\rebuttal{and F5M95} and \MBD), the Brier score for males is \rebuttal{strong} whereas that for females is \rebuttal{very weak}. When we train with a minority of male patients (\rebuttal{F95M5 and} \FBD), the opposite occurs (see Table \ref{tab:sex_pred_analysis_reinforcing}).

\subsection{Age-prediction head evaluation}
\begin{figure*}[htbp]
    \centering
    \begin{subfigure}[b]{0.86\textwidth}
        \includegraphics[width=\textwidth]{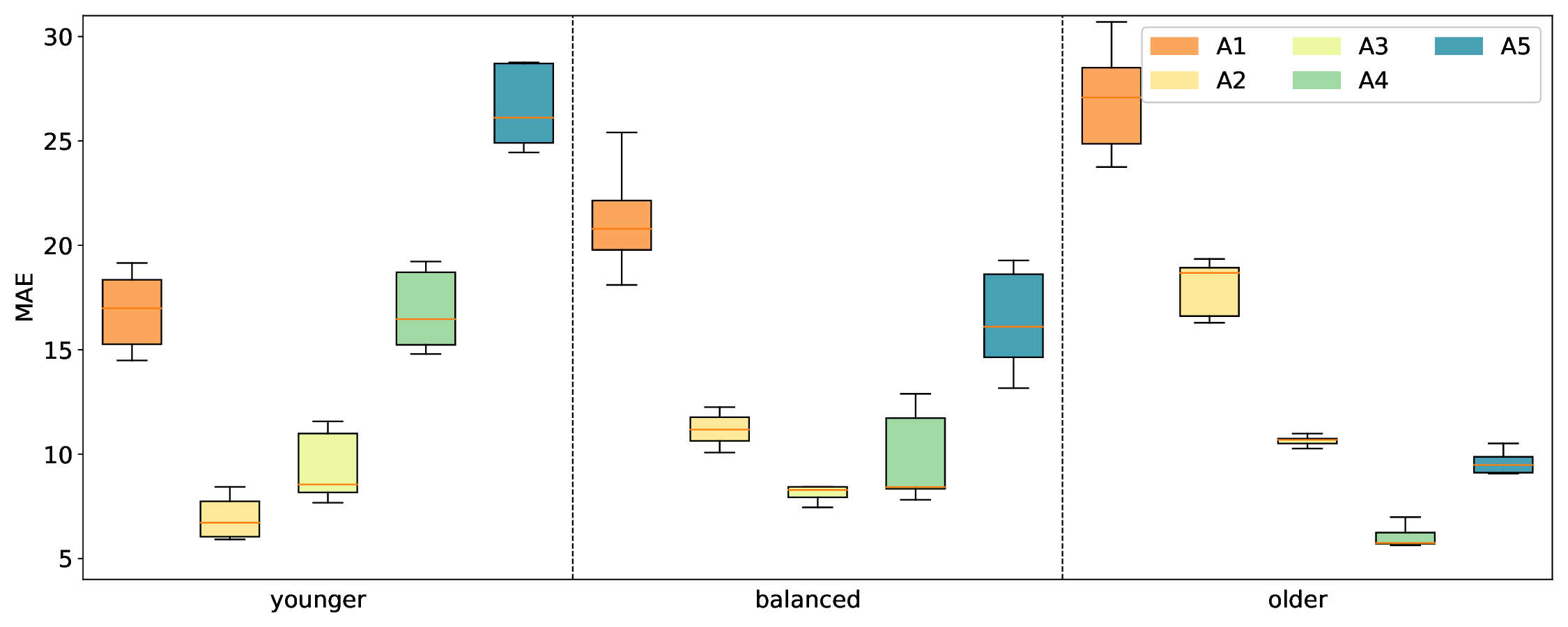}
    \end{subfigure}
    
    \vspace{1em}
    \hspace{0.5em}
    \begin{subfigure}[b]{0.25\textwidth}
        \includegraphics[width=\textwidth]{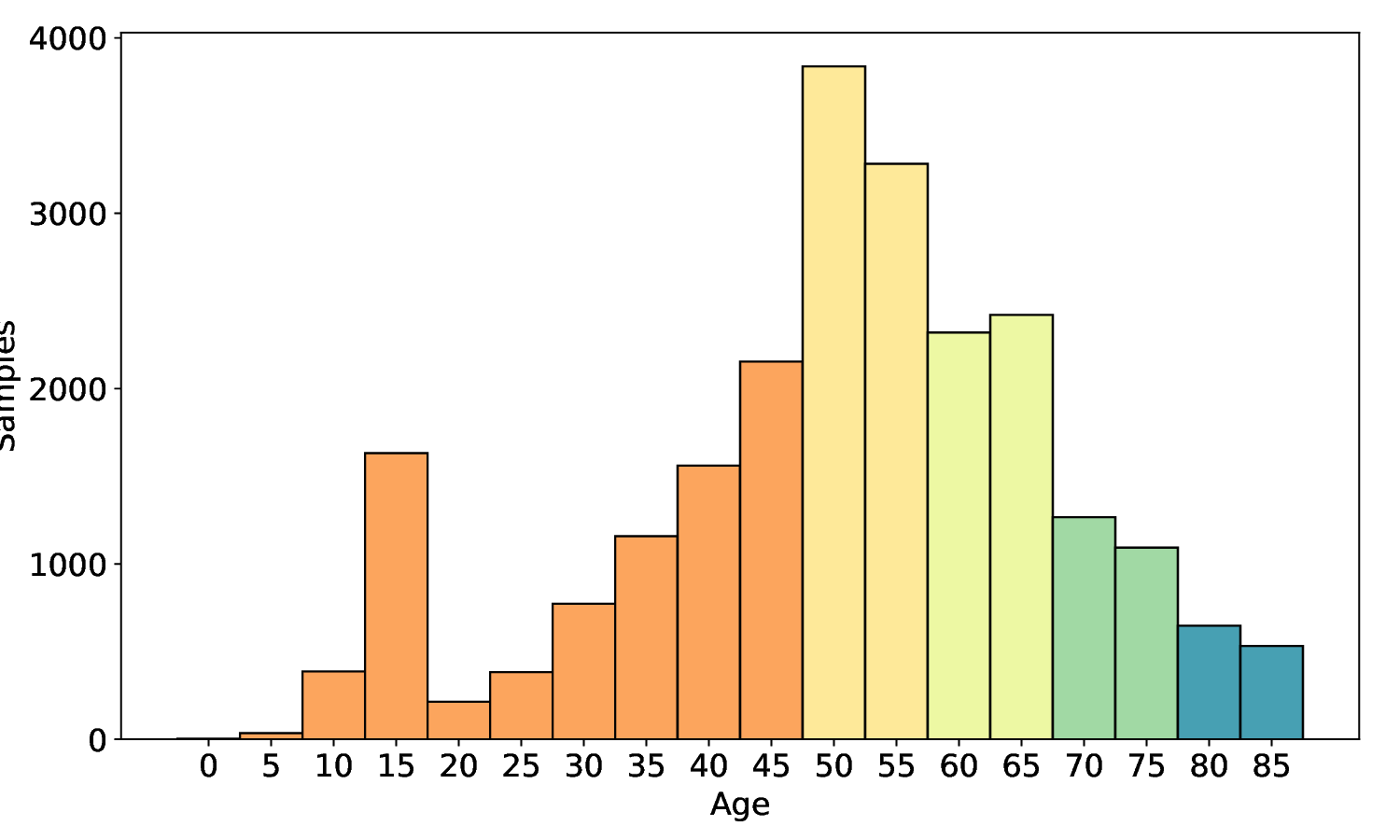}
    \end{subfigure}
    \hspace{0.5em}
    \begin{subfigure}[b]{0.25\textwidth}
        \includegraphics[width=\textwidth]{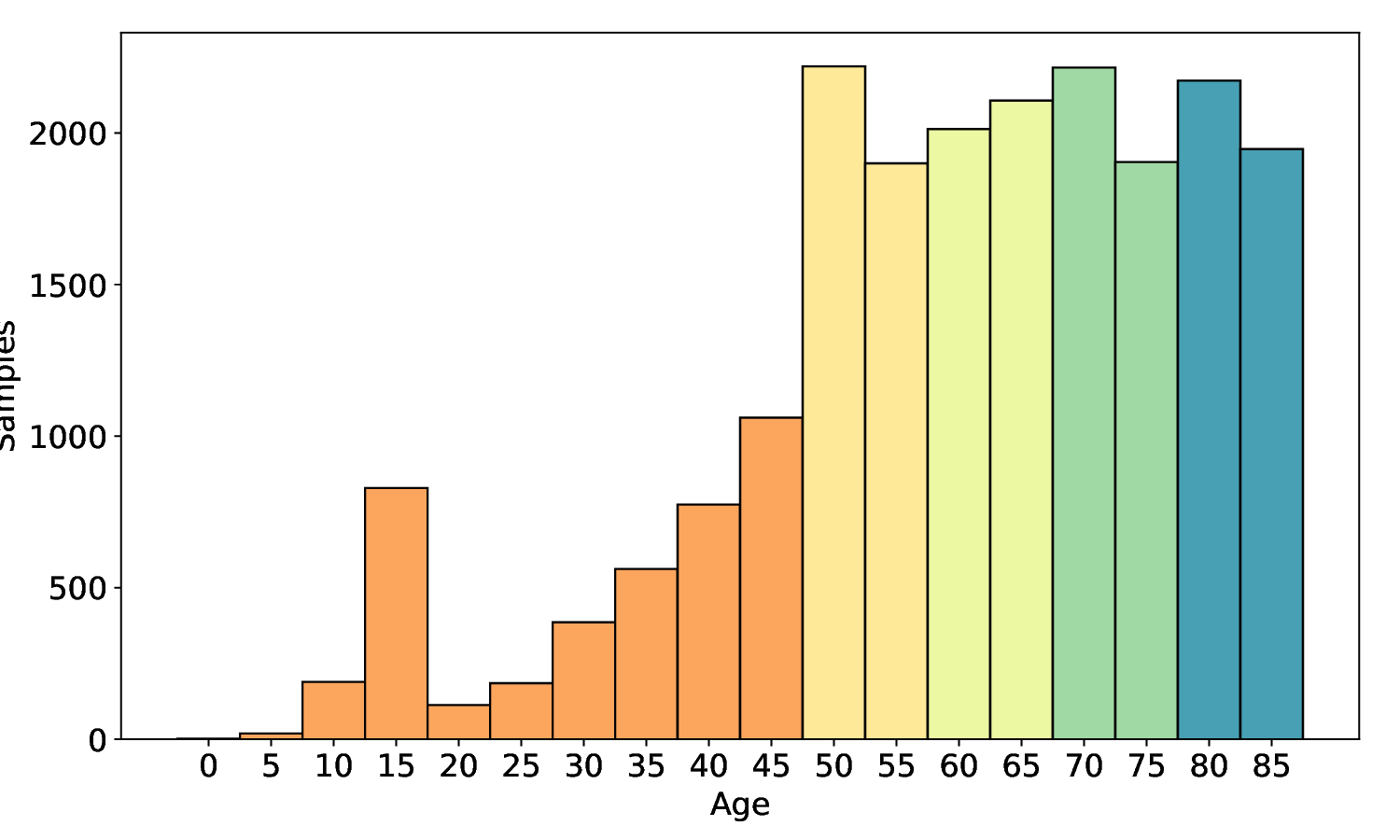}
    \end{subfigure}
    \hspace{0.5em}
    \begin{subfigure}[b]{0.25\textwidth}
        \includegraphics[width=\textwidth]{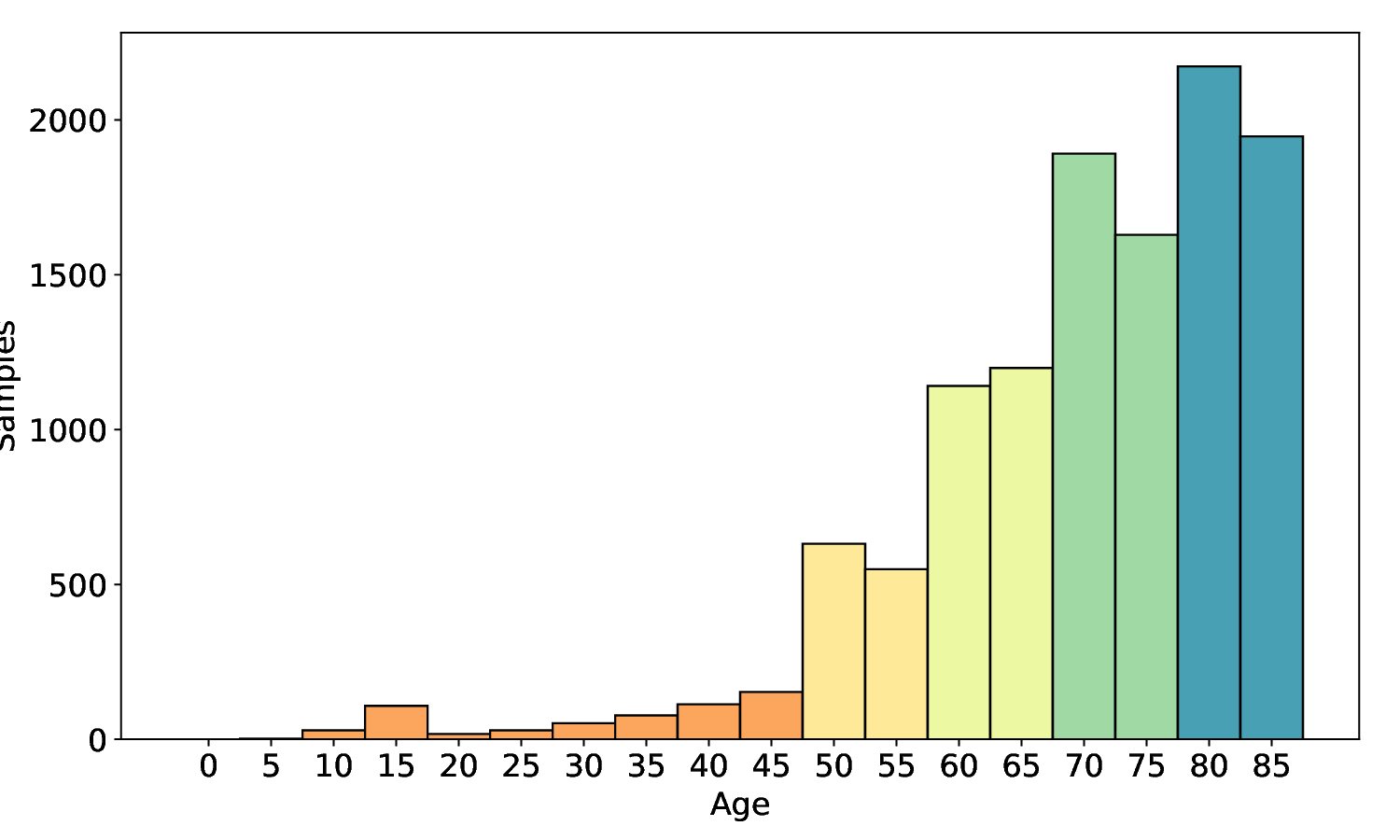}
    \end{subfigure}
    \caption{The top figure shows the Mean Absolute Error (MAE) distributions for five age groups ($A_{1}$–$A_{5}$). Each plot depicts the MAE, calculated from predicted versus reference ages, under one of the three training biases (younger, balanced, older) using the reinforcing model. We evaluated all models on a balanced test set. The bottom three figures show the age distribution for the younger, balanced, and older training datasets (from left to right). The same colour legend that appears in the top panel (mapping $A_{1}$–$A_{5}$ to their respective colours) is reused for the three lower panels. The age brackets are defined as follows, \rebuttal{where $a$ represents the patient's age in years: $A_1 = {0 \leq a \leq 50}$, $A_2 = {51 \leq a \leq 60}$, $A_3 = {61 \leq a \leq 70}$, $A_4 = {71 \leq a \leq 80}$, and $A_5 = {a \geq 81}$.}}
    \label{fig:reinforce_age_mae}
\end{figure*}

    We evaluated the auxiliary age prediction head by reporting both the Pearson correlation (Table \ref{tab:age_pearson}) and the distribution of the mean absolute error (Figure \ref{fig:reinforce_age_mae}). Table \ref{tab:age_MAE} provides the overall MAE for the three training-bias configurations.

\begin{table}[h]
\centering
\caption{Pearson correlation ($\rho$) between predicted and true ages for the reinforcing multi‑task model. Correlation coefficients are reported for three training‑bias configurations (younger‑skewed, balanced, older‑skewed) evaluated on a balanced test set. The table lists the overall $\rho$ as well as the $\rho$ for each age category. “-” indicates too low to be meaningfully reported.}

\small
\begin{tabular}{llll}
\hline
$\rho$ & YOUNGER& BALANCED& OLDER\\
\hline
overall& 0.319& 0.433& \textbf{0.517}\\
$A_{1}$& 0.354& 0.471& \textbf{0.710}\\
$A_{2}$& 0.104& 0.105& 0.078\\
$A_{3}$& -& 0.070& -\\
$A_{4}$& -& -& -\\
$A_{5}$& -& 0.069& -\\
\hline
\end{tabular}
\label{tab:age_pearson}
\end{table}

\rebuttal{In the BALANCED training configuration, the auxiliary head shows a moderate Pearson correlation ($\rho=0.433$), indicating that the model recognises age patterns without relying on them as a dominant shortcut. This supports the hypothesis that balanced sampling forces the encoder to learn features that generalise across age groups. In contrast, the older-skewed training yields a strong correlation ($\rho=0.710$) for the youngest cohort, suggesting the model relies heavily on shortcut-related age cues when the distribution is imbalanced.}

\paragraph{Increasing correlation with older-skewed training}
    The reinforcement model trained on the OLDER dataset exhibits a strong Pearson correlation between predicted and ground‑truth ages. In contrast, models trained on the BALANCED and YOUNGER datasets show only moderate \rebuttal{and weak} correlations, with the BALANCED model achieving a slightly higher correlation than the YOUNGER model (Table \ref{tab:age_pearson}).

\paragraph{Youngest cohort shows strongest correlation}
    $A_1$ correlates most strongly with older‑skewed data (Table \ref{tab:age_pearson}).

\paragraph{Middle group shows weakest correlation}
    The middle age groups ($A_2\text{--}A_4$) correlations are uniformly weak and often marked \rebuttal{with a dash "-" in} (Table \ref{tab:age_pearson}).

\paragraph{Oldest cohort shows practically no correlation}
    Only the balanced scheme shows a very weak correlation \rebuttal{for the oldest cohort ($A_5$)}; for the other schemes, the correlation score is marked \rebuttal{with a dash "-" in} Table \ref{tab:age_pearson}.

\paragraph{MAE of age categories}
    \begin{table}[h]
\centering
\caption{Mean Absolute Error (MAE) of the auxiliary age‑prediction head for the reinforcing multi‑task model under three training‑bias configurations.}
\small
\begin{tabular}{llll}
\hline
& YOUNGER& BALANCED& OLDER\\
\hline
MAE& $15.335 \pm 0.643$& $\textbf{13.409} \pm \textbf{1.031}$& $14.252 \pm 0.586$\\
\hline
\end{tabular}
\label{tab:age_MAE}
\end{table}
    We find that the model we trained on a balanced age distribution yields the lowest overall average error (Table \ref{tab:age_MAE}), whereas the younger‑skewed and older‑skewed configurations show higher MAE.
    For the youngest cohort of patients, we see relatively high MAE scores for all three models (Figure \ref{fig:reinforce_age_mae}). The MAE scores of the youngest cohort increase as the number of patients in that cohort in the training set decreases. We observe the same pattern for the oldest patient cohort. However, in skewed training, the oldest cohort scores better when trained with mainly older patients than the youngest cohort trained with mostly younger patients. \rebuttal{The oldest age groups ($A_4$ and $A_5$) exhibit high MAE when the model is trained mostly on younger patients, while the youngest age groups ($A_1$ and $A_2$) show high MAE when trained mainly on older patients.} The $A_2$ age cohort achieves the lowest score when we train with predominantly younger patients. The scores of the $A_3$ cohort change little in all models. The MAE score of the $A_4$ cohort decreases as the proportion of $A_4$ patients relative to the total training population increases. The $A_4$ age cohort achieves the lowest score when we train the model with mostly older patients.

\subsection{Cross-dataset analysis}

    \subsubsection{PAD-UFES-20}
\begin{figure*}[!ht]
        \centering
        \includegraphics[width=1\textwidth] {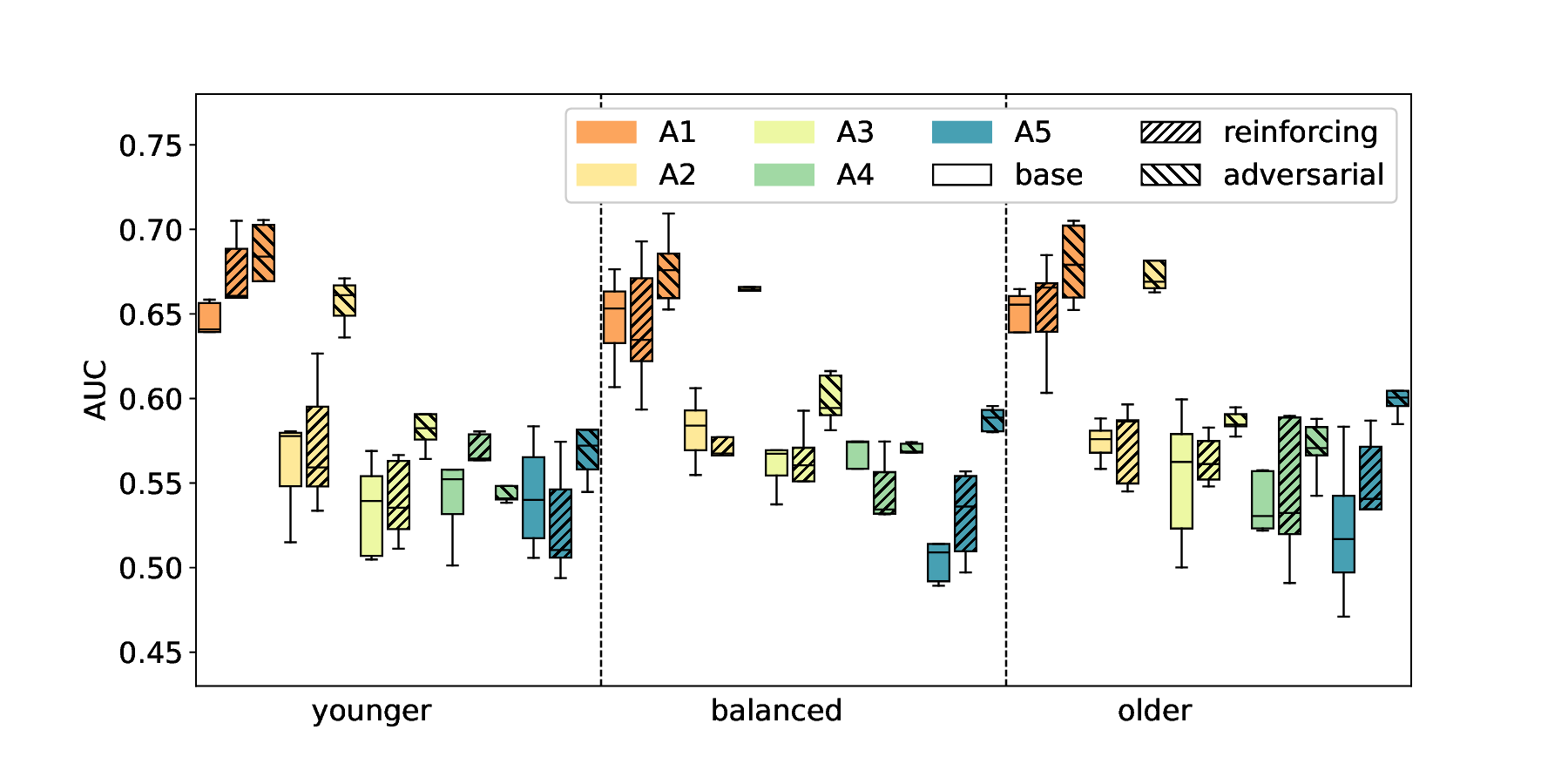}
        \caption{Age-stratified model evaluation showing AUC performance across different age categories (A1-A5) for three model architectures (base, reinforcing, and adversarial) trained and validated on age-biased ISIC datasets (YOUNGER, BALANCED, OLDER) and evaluated on the curated \textbf{PAD-UFES-20 dataset}. The analysis demonstrates how each model type performs across different age distributions. The age brackets are defined as follows, \rebuttal{where $a$ represents the patient's age in years: $A_1 = {0 \leq a \leq 50}$, $A_2 = {51 \leq a \leq 60}$, $A_3 = {61 \leq a \leq 70}$, $A_4 = {71 \leq a \leq 80}$, and $A_5 = {a \geq 81}$.}}
        \label{fig:auc_subhroups_age_padufes}
    \end{figure*}

\begin{figure*}[!ht]
        \centering
        \includegraphics[width=1\textwidth] {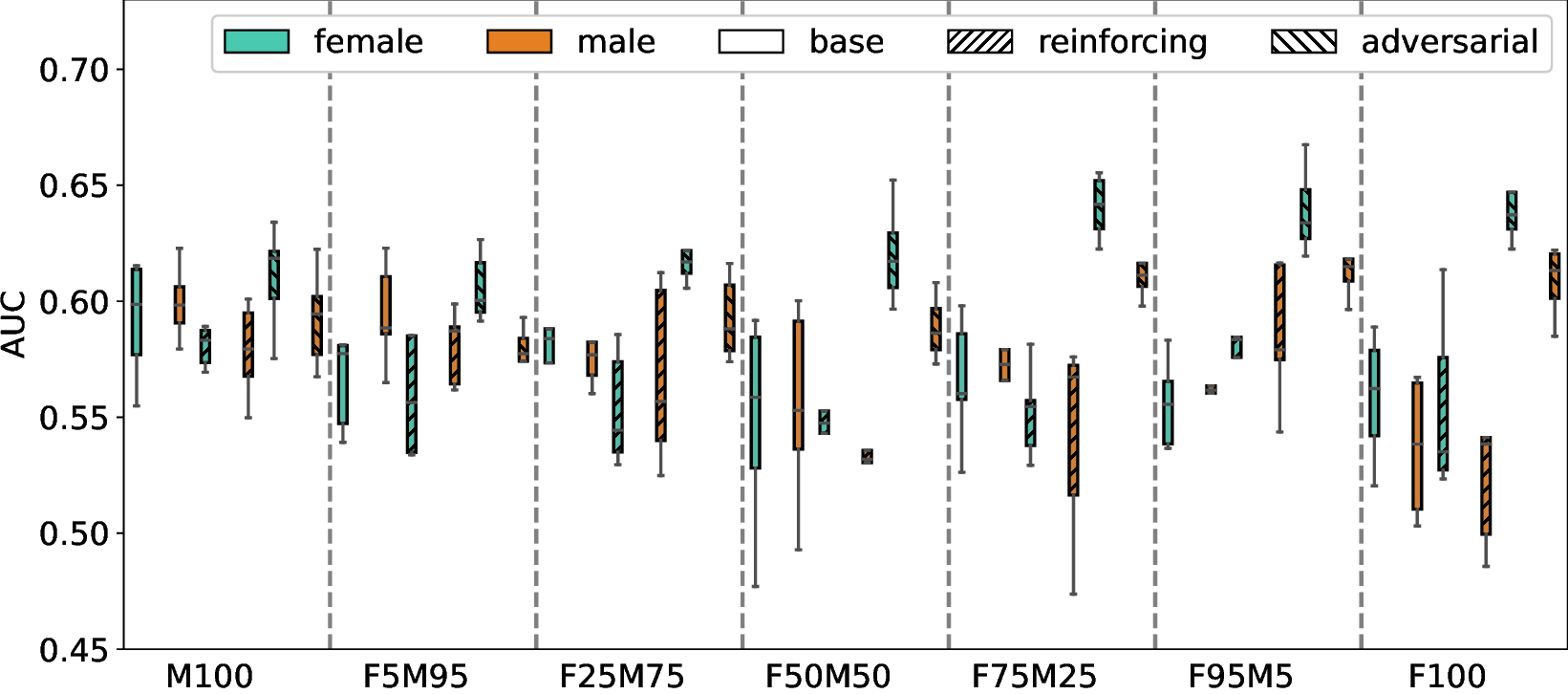}
        \caption{Sex-stratified model evaluation showing AUC performance across male and female categories for three model architectures (base, reinforcing, and adversarial) trained and validated on sex-biased ISIC datasets (\rebuttal{M100, F5M95, F25M75, F50M50, F75M25, F95M5 and F100}) and evaluated on the curated \textbf{PAD-UFES-20 dataset}.}
        \label{fig:auc_subhroups_sex_padufes}
    \end{figure*}

    Figures \ref{fig:auc_subhroups_age_padufes} and \ref{fig:auc_subhroups_sex_padufes} present our model performance evaluation on the PAD-UFES-20 dataset, examining age-stratified and sex-stratified distributions. While Figure \ref{fig:auc_subhroups_age_padufes} breaks down performance across five age categories (A1-A5) for different age-biased training sets, Figure \ref{fig:auc_subhroups_sex_padufes} analyses sex-based performance variations across multiple male-female distribution ratios.

    \paragraph{External validation shows performance drop}
        During external validation, model performance in both sex- and age-based evaluations showed notably lower metrics compared to internal validation scenarios (Figures \ref{fig:auc_subhroups_age_padufes} and \ref{fig:auc_subhroups_sex_padufes}).

    \paragraph{External validation shows best performance for younger age groups}
        For age-based models, we observe a different pattern in external validation compared to internal validation. The models trained on the YOUNGER, BALANCED, and OLDER cases show similar performance ranges in the three learning strategies for age categories $A_{2}$, $A_{3}$, $A_{4}$, and $A_{5}$. \rebuttal{Across all configurations, models show notably better performance for the} $A_{1}$ age category. The adversarial model also shows improved performance for the $A_{2}$ age category, achieving results comparable to the performance of the $A_{1}$ category (Figure \ref{fig:auc_subhroups_age_padufes}).

    \paragraph{Sex-based validations show contrasting patterns}
        In internal validation, we observed an X pattern: as the percentage of male patients in the training and validation sets decreased, male patients' performance declined, whereas female patients' performance improved. However, the X pattern is less pronounced than in the internal validation. For adversarial models, we observed a different pattern (see Figure \ref{fig:auc_subhroups_sex_padufes}): in the internal validation with F75M25 adversarial models, the performance for female patients was lower than for male patients, while in the internal validation with F100 adversarial models, the performance for male and female patients was equal. However, in external validation, we observe a notable improvement in performance in female patients.

    \subsubsection{DERM7PT}

\begin{figure*}[!ht]
        \centering
        \includegraphics[width=1\textwidth] {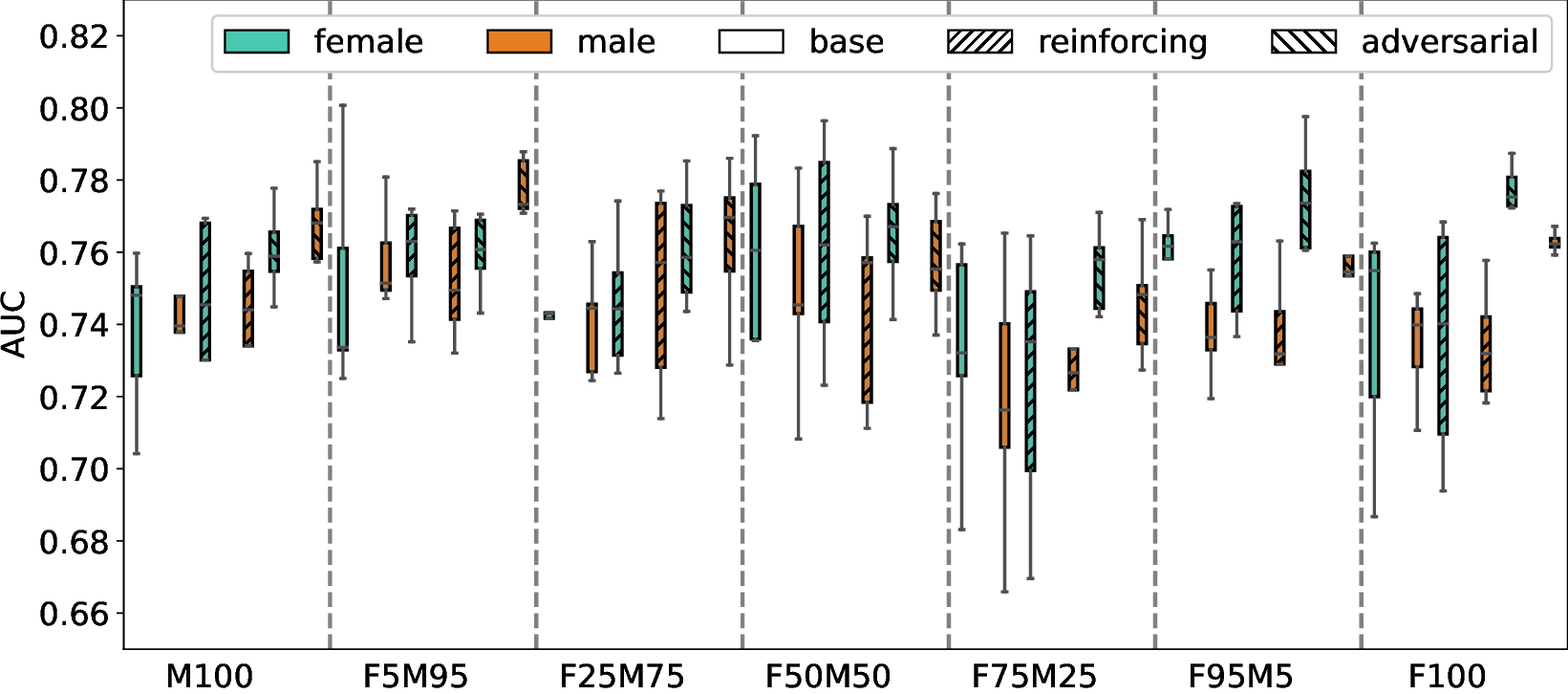}
        \caption{\rebuttal{Sex-stratified model evaluation showing AUC performance across male and female categories for three model architectures (base, reinforcing, and adversarial) trained and validated on sex-biased ISIC datasets (M100, F5M95, F25M75, F50M50, F75M25, F95M5 and F100) and evaluated on the curated \textbf{DERM7PT dataset}.}}
        \label{fig:auc_subhroups_sex_derm7pt}
    \end{figure*}

        \paragraph{\rebuttal{External validation shows performance drop}}
        \rebuttal{During external validation, we observed that the models showed considerably lower AUC metrics than in the internal validation experiments (see Figure \ref{fig:auc_subhroups_sex_derm7pt}). Nevertheless, we found that their performance remained higher than the sex‑based results obtained on PAD‑UFES‑20.}

        \paragraph{\rebuttal{Reduced subgroup performance gaps across external datasets}}
        \rebuttal{We observed that the performance of the male and female sub‑groups remained much closer together across the various training scenarios than it did in the ISIC‑based experiments, and less pronounced than that seen with PAD‑UFES‑20. In other words, the sex ratio distribution in the training data had a markedly smaller effect on the performance gaps between subgroups for the DERM7PT validation.}

\section{Discussion and conclusions}
    We investigated the effect of demographic bias on skin lesion classification performance using three ResNet-50-based CNN models, with a specific focus on variations in patient sex and age in the training data. \rebuttal{Using linear programming to generate datasets with controlled demographic distributions, we evaluated three learning strategies: a single-task model, a reinforcing multi-task model, and an adversarial learning scheme.} Additionally, we performed cross-dataset validation to assess the model's generalisation capabilities. Overall, the results highlight that sex‑related performance gaps are largely driven by training set imbalances, whereas age-related declines persist even with balanced sampling, and that domain shifts (from dermoscopic to smartphone images) cause substantial drops in external validation accuracy. \rebuttal{Bias patterns were not consistent across all datasets.}

\paragraph{\rebuttal{Sex based analysis}}
In our sex-based analysis, we observed that sex-specific training data produced better results, though single-task models exhibited notable sex bias. The reinforcement approach was partially effective in reducing this bias. The adversarial model eliminated sex bias specifically in cases involving \rebuttal{predominantly female distributions (F95M5)}.
\rebuttal{An unexpected result emerged regarding sex-related bias in ISIC-based datasets. When the patient cohort was male-dominated, models achieved higher accuracy on male patients. Conversely, in female-dominated cohorts, models did not show a comparable boost for female patients. This asymmetry suggests that models tend to overfit to male-specific cues. In contrast to ISIC-based datasets, PAD-UFES-20 and DERM7PT showed that the adversarial model performed better on female patients than on male patients in female-dominated cohorts.}

\rebuttal{The auxiliary sex-prediction head heavily relies on data composition: it learns demographic signals under balanced distributions but struggles in skewed scenarios. This shows that the reinforcing model cannot build a robust discriminator when the minority class is underrepresented, leading to the collapse of the regularisation signal. Our Brier score analysis confirms this mechanism: in balanced datasets, the head makes well-calibrated predictions for both sexes, whereas in skewed datasets, it becomes over-confident for the majority and inaccurate for the minority. Ultimately, effective debiasing depends on the auxiliary head's ability to reliably learn the attribute. When data scarcity prevents this, the reinforcing model overfits to the dominant group, and the regularisation mechanism fails.}

Our findings on sex-related bias contradict the conclusion of Sies et al. that \emph{despite sex-related imbalances in open access training data, the diagnostic performance of the CNN tested showed no sex-related bias in the classification of skin lesions} (\cite{Sies2022-dp}). We hypothesise that dataset size contributes to these divergent findings. Whereas Sies et al. leveraged over 150,000 dermoscopic images, our smaller dataset may have rendered the CNN more vulnerable to sex-related biases. Larger datasets typically offer greater diversity across demographic groups, facilitating the learning of robust and generalisable features. Conversely, smaller datasets may amplify existing biases or result in overfitting to specific demographic characteristics.

As expected, the base model shows sensitivity for sex bias, likely driven by overfitting and various anatomical confounders in the training data. These include sex-specific variations in skin thickness, hair distribution, lesion location, and underlying vasculature, as well as differences in sun exposure patterns linked to sex-linked behavioural factors. While the reinforcing and adversarial models incorporate regularisation techniques to mitigate such bias, our experiments revealed limited success: bias correction was observed only in the adversarial model for female-only cohorts. This suggests that complex anatomical confounders continue to substantially influence model predictions in mixed-sex populations, resisting standard regularisation approaches.

\paragraph{\rebuttal{Age based analysis}}
In age-related experiments, we found comparable baseline performance across all three model approaches, with an evident decline in performance in older age categories. We observed a clear decline trend across age categories: the age group $A_{1}$ consistently achieved the highest performance, while subsequent age groups showed progressively lower scores, regardless of the training data distribution used (YOUNGER, BALANCED or OLDER).

\rebuttal{The moderate predictive power of the age head ($0.319 \leq \rho \leq 0.517$) suggests it may not strongly influence bias mitigation. As shown in Figure \ref{fig:auc_subhroups_age}, performance disparities remain evident across all training distributions. While the OLDER dataset shows smaller subgroup gaps than the YOUNGER and BALANCED cases, we cannot definitively attribute this to successful bias mitigation. Instead, it appears to result from the model exploiting age-related shortcuts. This is supported by the strong correlation for the youngest cohort ($A_1$, $\rho = 0.710$) under older-skewed training, suggesting the model may rely on age-specific cues.}

A plausible explanation for the trend of increasing overall correlation from the youngest cohort to the oldest cohort is that the model trained on a cohort dominated by older patients learns a more reliable mapping from image characteristics to chronological age because aging skin shows more significant structural alterations, such as wrinkles, skin texture degradation, increased density of pigmentary spots, and larger pigmentary spots (\cite{Flament2019-os}).

Based on Pearson correlation and MAE analyses, the auxiliary age-prediction head functions as a demographically sensitive regressor that performs optimally when age distributions are balanced and match the training data. Fairness-focused pipelines should therefore employ balanced age sampling or explicit re-weighting and augmentation to mitigate systematic bias. This is particularly evident in the MAE analysis, which shows substantially higher errors for underrepresented age groups.

Several factors likely contribute to the observed age-related performance differences. Natural changes in skin characteristics with age can influence lesion classification. Even with balanced sampling, intrinsic skin conditions may differ between age groups, potentially affecting model performance.

\rebuttal{Within the youngest age group ($A_1$), malignant cases show a concentration toward the end of the age range, close to the upper boundary of the group. This contrasts with benign lesions, which are well-represented throughout the group, including at the beginning. This internal skew results in an unequal distribution within the category: an overrepresentation of benign samples at the beginning of the group and malignant samples at the end. Such a subgroup distribution can influence model training and complicate the interpretation of the corresponding test results. While we maintained class balance across age distributions through our LP constraints, we recognise that the natural prevalence of malignancy varies by age in clinical practice.}

The consistent decline in performance for older age categories initially suggests a need for more balanced age representation in training data. However, since this pattern persists even with balanced training sets, it indicates that factors beyond mere data distribution, such as intrinsic biological or physiological differences, are driving these performance gaps.

The discretisation of continuous variables such as age presents methodological challenges, given that age is inherently continuous. While categorical variables like patient sex have natural groupings, age categorisation necessitates arbitrary cut-off points that may not correspond to biologically or clinically meaningful boundaries.

\rebuttal{Adversarial learning is designed to reduce reliance on confounding features by training a discriminator to predict protected attributes, while updating the main model to minimise the discriminator's prediction.} However, our results show that this approach succeeded only in certain data compositions, failing to mitigate bias in other scenarios. This discrepancy suggests that shortcut learning (the reliance on superficial correlations rather than clinically relevant features) remains a significant concern. For instance, body hair patterns, which dermatologists have noted can significantly interfere with dermatoscopic examinations (\cite{Fink2020-sf}), could be unintentionally used by the model as a proxy for sex or age classification, potentially affecting performance differences between male and female patients. Similarly, confounding factors such as skin colour and image artefacts may influence classification performance across demographic subgroups. Future research should systematically investigate these factors to determine whether adversarial learning can be adapted to address them more robustly across all demographic compositions. Furthermore, we will examine the bias mechanisms by analysing the auxiliary head's behaviour in the adversarial setting.

We demonstrate that skewed distributions in training data cause performance disparities across \rebuttal{sex} and age groups. Consequently, when using a dataset with a certain level of skewness, one possible mitigation strategy is to rebalance the data through augmentation. However, in many practical scenarios, acquiring additional instances is often infeasible in the short term. Under such constraints, introducing synthetically generated images may provide an alternative way to restore balance (\cite{Stanley2024-yw}, \cite{Kebaili2023-zg}). Nonetheless, it remains crucial to understand the root sources of bias, such as demographic representation, acquisition protocols, or annotation practices, so these factors can be explicitly considered during the synthesis process.

\paragraph{\rebuttal{Cross-dataset validation and future directions}}
In our cross-dataset validation, we observed performance patterns by demographics: younger age groups generally performed better but showed a notable decline in performance during external validation. In the external validation of models trained on varying female-to-male ratios, we observed improved performance for female patients.

External-validation performance may be affected by numerous factors, including differences in training data, imaging equipment (such as smartphones versus dermoscopes), population demographics, and image-collection protocols. The PAD-UFES-20 set comprises digital camera photographs that differ markedly from the dermoscopic images used for training. Because we evaluated the models without any fine-tuning, we observed a sharp drop in accuracy for the PAD-UFES-20 set. We assume that the lack of adaptation contributed to this decline, although other factors may also play a role. Previous research has provided valuable information in this area. \rebuttal{DERM7PT showed lower AUC than internal validation but outperformed PAD-UFES-20, with smaller male/female performance gaps than ISIC-based experiments. This likely reflects modality consistency (dermoscopic-to-dermoscopic transfer reduces domain shift) and potentially less salient sex-related visual cues in DERM7PT images compared to ISIC.}

\rebuttal{Previous research has provided valuable information in this area.} Bevan and Atapour-Abarghouei demonstrated improved generalisation through “unlearning” spurious variations in skin lesion imaging instruments (\cite{Bevan2023-kg}). Daneshjou and colleagues emphasised the need to address skin tone bias in dermatology AI systems before deploying them to diverse populations (\cite{Daneshjou2022-ei}). These studies confirm that performance decline results from differences in image acquisition methods and varying population characteristics between datasets.  Additional research is needed to investigate cross-dataset performance across different imaging modalities, such as comparing model robustness between dermoscopic, smartphone, and conventional digital camera images. \rebuttal{We recommend that future studies separate modality effects from demographic bias. One way is to assemble paired datasets where the same lesions are photographed with dermoscopes, smartphones, and regular digital cameras. These “multi‑view” benchmarks would let us measure performance loss caused by modality changes while keeping lesion identity and patient demographics constant.}




\paragraph{Concluding remarks}
In conclusion, our experiments show that imbalanced training data mainly cause sex-related performance differences in skin-lesion classification. Conversely, age-related performance gaps remain even when the training set is balanced. 

Our findings highlight the importance of understanding two key aspects when designing and implementing AI in medical imaging: explicit factors (such as imaging protocols, patient demographic data, data set distributions, and sampling methods) and implicit factors (such as geographic differences, biological variations, and demographic imbalances). Specifically, our research concludes that even with balanced training sets, performance disparities between demographic groups persisted, indicating that the relationship between these factors is complex. Previous studies have shown that bias in medical imaging arises from multiple interconnected factors (\cite{Zong2023-dy}). Disentangling how these factors reinforce or oppose each other, as well as their impact on the performance of medical image applications, remains a challenging area for further research.

We hope that the methodologies and insights developed through our research can serve as building blocks to create more sophisticated and equitable healthcare AI systems that better serve diverse patient populations.


\newpage
\acks{This work was supported by the Netherlands Organisation for Scientific Research, grant no. 023.014.010.}

%
\ethics{The work follows appropriate ethical standards in conducting research and writing the manuscript, following all applicable laws and regulations regarding treatment of animals or human subjects.}

\coi{The authors declare that they have no known competing financial interests or personal relationships that could have influenced the work reported in this paper.}



\data{The data used to create the various datasets in this study is publicly available through the ISIC (International Skin Imaging Collaboration) archive, the PAD-UFES-20 dataset, \rebuttal{and the DERM7PT dataset}. The code necessary for dataset creation and analysis is available via a public GitHub repository. Researchers interested in reproducing or building on this work can access the ISIC archive at \url{https://www.isic-archive.com}, the PAD-UFES-20 dataset at \url{https://data.mendeley.com/datasets/zr7vgbcyr2/1}, \rebuttal{the DERM7PT dataset at} \url{http://derm.cs.sfu.ca}, and find our code at \url{https://github.com/raumannsr/demographic-fairness-extended}.}

\clearpage
\bibliography{bibliography/refs_ralf,bibliography/refs_gerard,bibliography/refs_veronika}

@ARTICLE{Adeli2021-da,
title = "Representation Learning with Statistical Independence to Mitigate Bias",
author = "Adeli, Ehsan and Zhao, Qingyu and Pfefferbaum, Adolf and Sullivan, Edith V and Fei-Fei, Li and Niebles, Juan Carlos and Pohl, Kilian M",
journal = "IEEE Winter Conf Appl Comput Vis",
volume =  2021,
pages = "2512--2522",
month =  jan,
year =  2021,
language = "en"
}

@ARTICLE{Amrhein2019-hl,
  title = "Scientists rise up against statistical significance",
  author = "Amrhein, Valentin and Greenland, Sander and McShane, Blake",
  journal = "Nature",
  publisher = "Springer Science and Business Media LLC",
  volume =  567,
  number =  7748,
  pages = "305--307",
  month =  mar,
  year =  2019,
  language = "en"
}

@misc{argenziano2000interactive,
  author    = {Argenziano, Giuseppe and others},
  title     = {Interactive Atlas of Dermoscopy: A Tutorial},
  howpublished = {Book and CD-ROM},
  year      = {2000},
  publisher = {Edra Medical Publishing and New Media},
  address   = {Milan, Italy},
  isbn      = {88-86457-30-8}
}

@ARTICLE{Bencevic2024-pg,
title = "Understanding skin color bias in deep learning-based skin lesion segmentation",
author = "Ben{\v c}evi{\'c}, Marin and Habijan, Marija and Gali{\'c}, Irena and Babin, Danilo and Pi{\v z}urica, Aleksandra",
journal = "Comput. Methods Programs Biomed.",
volume =  245,
pages = "108044",
month =  mar,
year =  2024,
language = "en"
}

@ARTICLE{Bevan2023-kg,
title = "Skin Deep Unlearning: Artefact and Instrument Debiasing in the Context of Melanoma Classification",
author = "Bevan, Peter J and Atapour-Abarghouei, Amir",
month =  apr,
year =  2023,
journal={arXiv preprint arXiv:2109.09818}
}

@misc{Bissoto2020-hn,
      title={Debiasing Skin Lesion Datasets and Models? Not So Fast}, 
      author={Alceu Bissoto and Eduardo Valle and Sandra Avila},
      year={2020},
      eprint={2004.11457},
      archivePrefix={arXiv},
      primaryClass={cs.CV}
}

@INCOLLECTION{Caruana1993,
  title = "Multitask learning: A knowledge-based source of inductive bias",
  author = "Caruana, Richard A",
  booktitle = "Machine Learning Proceedings 1993",
  publisher = "Elsevier",
  pages = "41--48",
  year =  1993
}

@ARTICLE{Cassidy2022-wx,
title = "Analysis of the {ISIC} image datasets: Usage, benchmarks and recommendations",
author = "Cassidy, Bill and Kendrick, Connah and Brodzicki, Andrzej and Jaworek-Korjakowska, Joanna and Yap, Moi Hoon",
journal = "Med. Image Anal.",
volume =  75,
pages = "102305",
month =  jan,
year =  2022,
language = "en"
}

@misc{Codella2017-rd,
      title={Skin Lesion Analysis Toward Melanoma Detection: A Challenge at the 2017 International Symposium on Biomedical Imaging (ISBI), Hosted by the International Skin Imaging Collaboration (ISIC)}, 
      author={Noel C. F. Codella and David Gutman and M. Emre Celebi and Brian Helba and Michael A. Marchetti and Stephen W. Dusza and Aadi Kalloo and Konstantinos Liopyris and Nabin Mishra and Harald Kittler and Allan Halpern},
      year={2018},
      eprint={1710.05006},
      archivePrefix={arXiv},
      primaryClass={cs.CV}
}

@misc{Codella2019-cn,
      title={Skin Lesion Analysis Toward Melanoma Detection 2018: A Challenge Hosted by the International Skin Imaging Collaboration (ISIC)}, 
      author={Noel Codella and Veronica Rotemberg and Philipp Tschandl and M. Emre Celebi and Stephen Dusza and David Gutman and Brian Helba and Aadi Kalloo and Konstantinos Liopyris and Michael Marchetti and Harald Kittler and Allan Halpern},
      year={2019},
      eprint={1902.03368},
      archivePrefix={arXiv},
      primaryClass={cs.CV}
}

@misc{Combalia2019-jj,
      title={BCN20000: Dermoscopic Lesions in the Wild}, 
      author={Marc Combalia and Noel C. F. Codella and Veronica Rotemberg and Brian Helba and Veronica Vilaplana and Ofer Reiter and Cristina Carrera and Alicia Barreiro and Allan C. Halpern and Susana Puig and Josep Malvehy},
      year={2019},
      eprint={1908.02288},
      archivePrefix={arXiv},
      primaryClass={eess.IV}
}

@ARTICLE{Daneshjou2022-ei,
  title = "Disparities in dermatology {AI} performance on a diverse, curated clinical image set",
  author = "Daneshjou, Roxana and Vodrahalli, Kailas and Novoa, Roberto A and Jenkins, Melissa and Liang, Weixin and Rotemberg, Veronica and Ko, Justin and Swetter, Susan M and Bailey, Elizabeth E and Gevaert, Olivier and Mukherjee, Pritam and Phung, Michelle and Yekrang, Kiana and Fong, Bradley and Sahasrabudhe, Rachna and Allerup, Johan A C and Okata-Karigane, Utako and Zou, James and Chiou, Albert S",
  journal = "Sci. Adv.",
  publisher = "American Association for the Advancement of Science (AAAS)",
  volume =  8,
  number =  32,
  pages = "eabq6147",
  month =  aug,
  year =  2022,
  language = "en"
}

@ARTICLE{Esteva2017-hn,
title = "Dermatologist-level classification of skin cancer with deep neural networks",
author = "Esteva, Andre and Kuprel, Brett and Novoa, Roberto A and Ko, Justin and Swetter, Susan M and Blau, Helen M and Thrun, Sebastian",
journal = "Nature",
volume =  542,
number =  7639,
pages = "115--118",
month =  feb,
year =  2017,
language = "en"
}

@ARTICLE{Ehteshami_Bejnordi2017-gr,
  author    = {Bejnordi, Babak Ehteshami and Veta, Mitko and van Diest, Paul Johannes and van Ginneken, Bram and Karssemeijer, Nico and Litjens, Geert and van der Laak, Jeroen AWM and Hermsen, Meyke and Manson, Quirine F and Balkenhol, Maschenka and others},
  title     = {Diagnostic Assessment of Deep Learning Algorithms for Detection of Lymph Node Metastases in Women With Breast Cancer},
  journal   = {JAMA},
  year      = {2017},
  volume    = {318},
  number    = {22},
  pages     = {2199--2210},
  publisher = {American Medical Association},
}

@ARTICLE{Fink2020-sf,
  title = "Physicians' level of hindrance by body hair in dermatoscopy and clinical benefit of an automated hair removal algorithm",
  author = "Fink, Christine and Uhlmann, Lorenz and Vogt, Karsten and Schneiderbauer, Roland and Menzer, Christian and Toberer, Ferdinand and Schank, Timo E and Enk, Alexander and Haenssle, Holger A",
  journal = "J. Dtsch. Dermatol. Ges.",
  publisher = "Wiley",
  volume =  18,
  number =  1,
  pages = "27--32",
  month =  jan,
  year =  2020,
  language = "en"
}

@ARTICLE{Flament2019-os,
  title = "Clinical impacts of sun exposures on the faces and hands of Japanese women of different ages",
  author = "Flament, F and Velleman, D and Yamamoto, S and Nicolas, A and Udodaira, K and Yamamoto, S and Morimoto, C and Belkebla, S and Negre, C and Delaunay, C",
  journal = "Int. J. Cosmet. Sci.",
  publisher = "Wiley",
  volume =  41,
  number =  5,
  pages = "425--436",
  month =  oct,
  year =  2019,
  language = "en"
}

@ARTICLE{Geirhos2020-te,
title = "Shortcut learning in deep neural networks",
author = "Geirhos, Robert and Jacobsen, J{\"o}rn-Henrik and Michaelis, Claudio and Zemel, Richard and Brendel, Wieland and Bethge, Matthias and Wichmann, Felix A",
journal = "Nature Machine Intelligence",
publisher = "Nature Publishing Group",
volume =  2,
number =  11,
pages = "665--673",
month =  nov,
year =  2020,
language = "en"
}

@ARTICLE{Gerrits2021-yo,
  title = "Publisher Correction: Age and sex affect deep learning prediction of cardiometabolic risk factors from retinal images",
  author = "Gerrits, Nele and Elen, Bart and Van Craenendonck, Toon and Triantafyllidou, Danai and Petropoulos, Ioannis N and Malik, Rayaz A and De Boever, Patrick",
  journal = "Sci. Rep.",
  publisher = "Springer Science and Business Media LLC",
  volume =  11,
  number =  1,
  pages =  1198,
  month =  jan,
  year =  2021,
  language = "en"
}

@BOOK{Geron2022-bm,
title = "{Hands-On} Machine Learning with {Scikit-Learn}, Keras, and {TensorFlow}",
author = "G{\'e}ron, Aur{\'e}lien",
publisher = "``O'Reilly Media, Inc.''",
month =  oct,
year =  2022,
language = "en"
}

@ARTICLE{Gichoya2022-nn,
  title = "{AI} recognition of patient race in medical imaging: a modelling study",
  author = "Gichoya, Judy Wawira and Banerjee, Imon and Bhimireddy, Ananth Reddy and Burns, John L and Celi, Leo Anthony and Chen, Li-Ching and Correa, Ramon and Dullerud, Natalie and Ghassemi, Marzyeh and Huang, Shih-Cheng and Kuo, Po-Chih and Lungren, Matthew P and Palmer, Lyle J and Price, Brandon J and Purkayastha, Saptarshi and Pyrros, Ayis T and Oakden-Rayner, Lauren and Okechukwu, Chima and Seyyed-Kalantari, Laleh and Trivedi, Hari and Wang, Ryan and Zaiman, Zachary and Zhang, Haoran",
  journal = "Lancet Digit. Health",
  publisher = "Elsevier BV",
  volume =  4,
  number =  6,
  pages = "e406--e414",
  month =  jun,
  year =  2022,
  language = "en"
}

@ARTICLE{Glocker2023-db,
  title = "Risk of bias in Chest Radiography deep learning foundation models",
  author = "Glocker, Ben and Jones, Charles and Roschewitz, Mélanie and Winzeck, Stefan",
  journal = "Radiol. Artif. Intell.",
  publisher = "Radiological Society of North America (RSNA)",
  volume =  5,
  number =  6,
  pages = "e230060",
  month =  nov,
  year =  2023,
  language = "en"
}

@ARTICLE{Groh2021-vm,
title = "Evaluating deep neural networks trained on clinical images in dermatology with the Fitzpatrick 17k dataset",
author = "Groh, Matthew and Harris, Caleb and Soenksen, L and Lau, Felix and Han, Rachel and Kim, Aerin and Koochek, A and Badri, Omar",
journal = "2021 IEEE/CVF Conference on Computer Vision and Pattern Recognition Workshops (CVPRW)",
pages = "1820--1828",
month =  apr,
year =  2021
}

@ARTICLE{Groh2022-ja,
title = "Towards Transparency in Dermatology Image Datasets with Skin Tone Annotations by Experts, Crowds, and an Algorithm",
author = "Groh, Matthew and Harris, Caleb and Daneshjou, Roxana and Badri, Omar and Koochek, Arash",
journal = "Proc. ACM Hum.-Comput. Interact.",
publisher = "Association for Computing Machinery",
volume =  6,
number = "CSCW2",
pages = "1--26",
month =  nov,
year =  2022,
address = "New York, NY, USA"
}

@misc{Gutman2016-lf,
      title={Skin Lesion Analysis toward Melanoma Detection: A Challenge at the International Symposium on Biomedical Imaging (ISBI) 2016, hosted by the International Skin Imaging Collaboration (ISIC)}, 
      author={David Gutman and Noel C. F. Codella and Emre Celebi and Brian Helba and Michael Marchetti and Nabin Mishra and Allan Halpern},
      year={2016},
      eprint={1605.01397},
      archivePrefix={arXiv},
      primaryClass={cs.CV}
}

@INPROCEEDINGS{He2016-ow,
title = "Deep residual learning for image recognition",
booktitle = "Proceedings of the {IEEE} conference on computer vision and pattern recognition",
author = "He, Kaiming and Zhang, Xiangyu and Ren, Shaoqing and Sun, Jian",
pages = "770--778",
year =  2016
}

@MISC{Isic-query,
title = "{ISIC} Archive",
howpublished = "\url{https://gallery.isic-archive.com}",
note = "Accessed: 2024-06-07",
key = "ISIC2024"
}

@ARTICLE{Jones2025-zn,
  title = "A primer on causal and statistical dataset biases for fair and robust image analysis",
  author = "Jones, Charles and Glocker, Ben",
  journal = "arXiv [cs.LG]",
  year =  2025,
  primaryClass = "cs.LG"
}

@ARTICLE{Kawahara2018-tx,
  title = "7-point checklist and skin lesion classification using multi-task multi-modal neural nets",
  author = "Kawahara, Jeremy and Daneshvar, Sara and Argenziano, Giuseppe and Hamarneh, Ghassan",
  journal = "IEEE J. Biomed. Health Inform.",
  publisher = "Institute of Electrical and Electronics Engineers (IEEE)",
  volume =  23,
  number =  2,
  pages = "538--546",
  month =  apr,
  year =  2018,
  language = "en"
}

@ARTICLE{Kebaili2023-zg,
  title = "Deep learning approaches for data augmentation in medical imaging: A review",
  author = "Kebaili, Aghiles and Lapuyade-Lahorgue, Jérôme and Ruan, Su",
  journal = "J. Imaging",
  publisher = "MDPI AG",
  volume =  9,
  number =  4,
  pages =  81,
  month =  apr,
  year =  2023,
  language = "en"
}

@ARTICLE{Ktena2024-bt,
  title = "Generative models improve fairness of medical classifiers under distribution shifts",
  author = "Ktena, Ira and Wiles, Olivia and Albuquerque, Isabela and Rebuffi, Sylvestre-Alvise and Tanno, Ryutaro and Roy, Abhijit Guha and Azizi, Shekoofeh and Belgrave, Danielle and Kohli, Pushmeet and Cemgil, Taylan and Karthikesalingam, Alan and Gowal, Sven",
  journal = "Nat. Med.",
  publisher = "Springer Science and Business Media LLC",
  volume =  30,
  number =  4,
  pages = "1166--1173",
  month =  apr,
  year =  2024,
  language = "en"
}

@ARTICLE{Kittler2002-kl,
  title = "Diagnostic accuracy of dermoscopy",
  author = "Kittler, H and Pehamberger, H and Wolff, K and Binder, M",
  journal = "Lancet Oncol.",
  publisher = "Elsevier BV",
  volume =  3,
  number =  3,
  pages = "159--165",
  month =  mar,
  year =  2002,
  language = "en"
}

@ARTICLE{Klingenberg2023-rk,
  title = "Higher performance for women than men in {MRI}-based Alzheimer's disease detection",
  author = "Klingenberg, Malte and Stark, Didem and Eitel, Fabian and Budding, Céline and Habes, Mohamad and Ritter, Kerstin and {Alzheimer's Disease Neuroimaging Initiative}",
  journal = "Alzheimers. Res. Ther.",
  publisher = "Springer Science and Business Media LLC",
  volume =  15,
  number =  1,
  pages =  84,
  month =  apr,
  year =  2023,
  language = "en"
}

@ARTICLE{liu2019-gz,
  title = "Joint classification and regression via deep multi-task multi-channel learning for Alzheimer's disease diagnosis",
  author = "Liu, Mingxia and Zhang, Jun and Adeli, Ehsan and Shen, Dinggang",
  journal = "IEEE Trans. Biomed. Eng.",
  publisher = "Institute of Electrical and Electronics Engineers (IEEE)",
  volume =  66,
  number =  5,
  pages = "1195--1206",
  month =  may,
  year =  2019,
  language = "en"
}

@ARTICLE{Nauta2021-fe,
title = "Uncovering and Correcting Shortcut Learning in Machine Learning Models for Skin Cancer Diagnosis",
author = "Nauta, Meike and Walsh, Ricky and Dubowski, Adam and Seifert, Christin",
journal = "Diagnostics (Basel)",
volume =  12,
number =  1,
month =  dec,
year =  2021,
language = "en"
}

@ARTICLE{Paszke1912-ny,
title = "An imperative style, high-performance deep learning library",
author = "Paszke, A and Gross, S and Massa, F and Lerer, A and Bradbury, J P and Chanan, G and Killeen, T and Lin, Z and Gimelshein, N and Antiga, L and {Others}",
journal = "Adv. Neural Inf. Process. Syst.",
volume =  32,
pages = "8026-8037",
year =  2019
}

@InProceedings{Raumanns2025-nt,
author="Raumanns, Ralf
and Schouten, Gerard
and Pluim, Josien P. W.
and Cheplygina, Veronika",
editor="Puyol-Ant{\'o}n, Esther
and Zamzmi, Ghada
and Feragen, Aasa
and King, Andrew P.
and Cheplygina, Veronika
and Ganz-Benjaminsen, Melanie
and Ferrante, Enzo
and Glocker, Ben
and Petersen, Eike
and Baxter, John S. H.
and Rekik, Islem
and Eagleson, Roy",
title="Dataset Distribution Impacts Model Fairness: Single Vs. Multi-task Learning",
booktitle="Ethics and Fairness in Medical Imaging",
year="2025",
publisher="Springer Nature Switzerland",
address="Cham",
pages="14--23",
}

@ARTICLE{Rufibach2010-tb,
  title = "Use of Brier score to assess binary predictions",
  author = "Rufibach, Kaspar",
  journal = "J. Clin. Epidemiol.",
  publisher = "Elsevier BV",
  volume =  63,
  number =  8,
  pages = "938--9; author reply 939",
  month =  aug,
  year =  2010,
  language = "en"
}

@ARTICLE{Saha2024-nq,
title = "Artificial intelligence and radiologists in prostate cancer detection on MRI (PI-CAI): an international, paired, non-inferiority, confirmatory study",
author = "Saha, A. and Bosma, J.S. and Twilt, J.J. and van Ginneken, B. and Bjartell, A. and Padhani, A.R. and Bonekamp, D. and Villeirs, G. and Salomon, G. and Giannarini, G. and Kalpathy-Cramer, J. and Barentsz, J. and Maier-Hein, K.H. and Rusu, M. and Rouvière, O. and van den Bergh, R. and Panebianco, V. and Kasivisvanathan, V. and Obuchowski, N.A. and Yakar, D. and others",
journal = "Lancet Oncol.",
month = jun,
year = 2024
}

@INPROCEEDINGS{Seth2024-ai,
title = "Does the Fairness of Your {Pre-Training} Hold Up? Examining the Influence of {Pre-Training} Techniques on Skin Tone Bias in Skin Lesion Classification",
booktitle = "Proceedings of the {IEEE/CVF} Winter Conference on Applications of Computer Vision",
author = "Seth, Pratinav and Pai, Abhilash K",
pages = "570--577",
year =  2024
}

@ARTICLE{Seyyed-Kalantari2021-dv,
  title = "Underdiagnosis bias of artificial intelligence algorithms applied to chest radiographs in under-served patient populations",
  author = "Seyyed-Kalantari, Laleh and Zhang, Haoran and McDermott, Matthew B A and Chen, Irene Y and Ghassemi, Marzyeh",
  journal = "Nat. Med.",
  publisher = "Springer Science and Business Media LLC",
  volume =  27,
  number =  12,
  pages = "2176--2182",
  month =  dec,
  year =  2021,
  language = "en"
}

@ARTICLE{Sies2022-dp,
  title = "Does sex matter? Analysis of sex-related differences in the diagnostic performance of a market-approved convolutional neural network for skin cancer detection",
  author = "Sies, Katharina and Winkler, Julia K and Fink, Christine and Bardehle, Felicitas and Toberer, Ferdinand and Buhl, Timo and Enk, Alexander and Blum, Andreas and Stolz, Wilhelm and Rosenberger, Albert and Haenssle, Holger A",
  journal = "Eur. J. Cancer",
  publisher = "Elsevier BV",
  volume =  164,
  pages = "88--94",
  month =  mar,
  year =  2022,
  language = "en"
}

@ARTICLE{Stanley2024-yw,
  title = "Towards objective and systematic evaluation of bias in artificial intelligence for medical imaging",
  author = "Stanley, Emma A M and Souza, Raissa and Winder, Anthony J and Gulve, Vedant and Amador, Kimberly and Wilms, Matthias and Forkert, Nils D",
  journal = "J. Am. Med. Inform. Assoc.",
  volume =  31,
  number =  11,
  pages = "2613--2621",
  month =  nov,
  year =  2024,
  language = "en"
}

@MISC{Tschandl2018-sz,
title = "The {HAM10000} dataset, a large collection of multi-source dermatoscopic images of common pigmented skin lesions",
author = "Tschandl, Philipp and Rosendahl, Cliff and Kittler, Harald",
journal = "Scientific Data",
volume =  5,
number =  1,
year =  2018
}

@ARTICLE{Vaidya2024-xz,
  title = "Demographic bias in misdiagnosis by computational pathology models",
  author = "Vaidya, Anurag and Chen, Richard J and Williamson, Drew F K and Song, Andrew H and Jaume, Guillaume and Yang, Yuzhe and Hartvigsen, Thomas and Dyer, Emma C and Lu, Ming Y and Lipkova, Jana and Shaban, Muhammad and Chen, Tiffany Y and Mahmood, Faisal",
  journal = "Nat. Med.",
  publisher = "Springer Science and Business Media LLC",
  volume =  30,
  number =  4,
  pages = "1174--1190",
  month =  apr,
  year =  2024,
  language = "en"
}

@ARTICLE{Veronica2021patient,
title = "A patient-centric dataset of images and metadata for identifying melanomas using clinical context",
author = "Rotemberg, V. and Kurtansky, N. and Betz-Stablein, B. and Caffery, L. and Chousakos, E. and Codella, N. and Combalia, M. and Dusza, S. and Guitera, P. and Gutman, D. and Halpern, A. and Helba, B. and Kittler, H. and Kose, K. and Langer, S. and Lioprys, K. and Malvehy, J. and Musthaq, S. and Nanda, J. and Reiter, O. and Shih, G. and Stratigos, A. and Tschandl, P. and Weber, J. and Soyer, H.P.",
journal = "Scientific Data; London",
publisher = "Nature Publishing Group",
volume =  8,
number =  1,
pages = "s41597--021",
year =  2021,
address = "London, United States, London",
language = "en"
}

@INPROCEEDINGS{Wu2022-hl,
title = "{FairPrune}: Achieving Fairness Through Pruning for Dermatological Disease Diagnosis",
booktitle = "Medical Image Computing and Computer Assisted Intervention -- {MICCAI} 2022",
author = "Wu, Yawen and Zeng, Dewen and Xu, Xiaowei and Shi, Yiyu and Hu, Jingtong",
publisher = "Springer Nature Switzerland",
pages = "743--753",
year =  2022
}

@ARTICLE{Xu2023-cg,
  title = "{ResNet} and its application to medical image processing: Research progress and challenges",
  author = "Xu, Wanni and Fu, You-Lei and Zhu, Dongmei",
  journal = "Comput. Methods Programs Biomed.",
  publisher = "Elsevier BV",
  volume =  240,
  number =  107660,
  pages =  107660,
  month =  oct,
  year =  2023,
  language = "en"
}

@ARTICLE{Yang2023-cw,
title = "An adversarial training framework for mitigating algorithmic biases in clinical machine learning",
author = "Yang, Jenny and Soltan, Andrew A S and Eyre, David W and Yang, Yang and Clifton, David A",
journal = "NPJ Digit Med",
volume =  6,
number =  1,
pages = "55",
month =  mar,
year =  2023,
language = "en"
}

@ARTICLE{Yi2021-nm,
  title = "Radiology ``forensics'': determination of age and sex from chest radiographs using deep learning",
  author = "Yi, Paul H and Wei, Jinchi and Kim, Tae Kyung and Shin, Jiwon and Sair, Haris I and Hui, Ferdinand K and Hager, Gregory D and Lin, Cheng Ting",
  journal = "Emerg. Radiol.",
  publisher = "Springer Science and Business Media LLC",
  volume =  28,
  number =  5,
  pages = "949--954",
  month =  oct,
  year =  2021,
  language = "en"
}

@ARTICLE{Zhang2022,
  title = "A Survey on Multi-Task Learning",
  author = "Zhang, Yu and Yang, Qiang",
  journal = "IEEE Trans. Knowl. Data Eng.",
  publisher = "Institute of Electrical and Electronics Engineers (IEEE)",
  volume =  34,
  number =  12,
  pages = "5586--5609",
  month =  dec,
  year =  2022
}

@ARTICLE{Zong2023-dy,
  title = "{MEDFAIR}: Benchmarking fairness for medical imaging",
  author = "Zong, Yongshuo and Yang, Yongxin and Hospedales, Timothy",
  journal = "arXiv [cs.LG]",
  month =  feb,
  year =  2023,
  archivePrefix = "arXiv",
  primaryClass = "cs.LG"
}

@Article{ruder2017overview,
  author  = {Ruder, Sebastian},
  title   = {An overview of multi-task learning in deep neural networks},
  journal = {arXiv preprint arXiv:1706.05098},
  year    = {2017},
}

@Article{willemink2020preparing,
  author    = {Willemink, Martin J and Koszek, Wojciech A and Hardell, Cailin and Wu, Jie and Fleischmann, Dominik and Harvey, Hugh and Folio, Les R and Summers, Ronald M and Rubin, Daniel L and Lungren, Matthew P},
  title     = {Preparing Medical Imaging Data for Machine Learning},
  journal   = {Radiology},
  year      = {2020},
  pages     = {192224},
  publisher = {Radiological Society of North America},
}

@InProceedings{abbasi2020risk,
  author    = {Abbasi-Sureshjani, Samaneh and Raumanns, Ralf and Michels, Britt EJ and Schouten, Gerard and Cheplygina, Veronika},
  booktitle = {MICCAI LABELS workshop, Lecture Notes in Computer Science},
  title     = {Risk of Training Diagnostic Algorithms on Data with Demographic Bias},
  year      = {2020},
  pages     = {183--192},
  publisher = {Springer},
  volume    = {12446},
  longvenue = {Interpretable and Annotation-Efficient Learning for Medical Image Computing: Third International Workshop, iMIMIC 2020, Second International Workshop, MIL3ID 2020, and 5th International Workshop, LABELS 2020, Held in Conjunction with MICCAI 2020, Lima, Peru, October 4–8, 2020, Proceedings 3 (pp. 183-192)},
}

@Article{larrazabal2020gender,
  author    = {Larrazabal, Agostina J and Nieto, Nicol{\'a}s and Peterson, Victoria and Milone, Diego H and Ferrante, Enzo},
  journal   = {Proceedings of the National Academy of Sciences},
  title     = {Gender imbalance in medical imaging datasets produces biased classifiers for computer-aided diagnosis},
  year      = {2020},
  number    = {23},
  pages     = {12592--12594},
  volume    = {117},
  publisher = {National Acad Sciences},
}

@Article{gichoya2022ai,
  author    = {Gichoya, Judy Wawira and Banerjee, Imon and Bhimireddy, Ananth Reddy and Burns, John L and Celi, Leo Anthony and Chen, Li-Ching and Correa, Ramon and Dullerud, Natalie and Ghassemi, Marzyeh and Huang, Shih-Cheng and others},
  journal   = {The Lancet Digital Health},
  title     = {{AI} recognition of patient race in medical imaging: a modelling study},
  year      = {2022},
  number    = {6},
  pages     = {e406--e414},
  volume    = {4},
  publisher = {Elsevier},
}

@Article{pacheco2020pad,
  author    = {Pacheco, Andre GC and Lima, Gustavo R and Salomao, Amanda S and Krohling, Breno and Biral, Igor P and de Angelo, Gabriel G and Alves Jr, F{\'a}bio CR and Esgario, Jos{\'e} GM and Simora, Alana C and Castro, Pedro BC and others},
  journal   = {Data in brief},
  title     = {PAD-UFES-20: A skin lesion dataset composed of patient data and clinical images collected from smartphones},
  year      = {2020},
  pages     = {106221},
  volume    = {32},
  publisher = {Elsevier},
}

@InProceedings{jimenez2023detecting,
  author       = {Jim{\'e}nez-S{\'a}nchez, Amelia and Juodelyte, Dovile and Chamberlain, Bethany and Cheplygina, Veronika},
  booktitle    = {2023 IEEE 20th International Symposium on Biomedical Imaging (ISBI)},
  title        = {Detecting shortcuts in medical images-a case study in chest x-rays},
  year         = {2023},
  organization = {IEEE},
  pages        = {1--5},
}

@InProceedings{petersen2022feature,
  author       = {Petersen, Eike and Feragen, Aasa and da Costa Zemsch, Maria Luise and Henriksen, Anders and Wiese Christensen, Oskar Eiler and Ganz, Melanie and Alzheimer’s Disease Neuroimaging Initiative},
  booktitle    = {International Conference on Medical Image Computing and Computer-Assisted Intervention},
  title        = {Feature robustness and sex differences in medical imaging: a case study in MRI-based Alzheimer’s disease detection},
  year         = {2022},
  organization = {Springer},
  pages        = {88--98},
}

\clearpage
\appendix
    \section{Sex distribution LP model}\label{lp_model_sex_dist}
    \subsection{Decision variables}
    \begin{minipage}{\columnwidth}
\setlength{\parskip}{0pt}
\setlength{\parindent}{0pt}
\setlength{\itemsep}{0pt}
\small
\begin{tabular}{@{}r@{\hspace{4pt}}l@{}}
    $x_{1}$: & \# malignant instances \\
    $x_{2}$: & \# benign instances \\
    $x_{3}$: & \# male patients (M) with malignant lesions \\
    $x_{4}$: & \# female patients (F) with malignant lesions \\
    $x_{5}$: & \# benign M \\
    $x_{6}$: & \# benign F \\
    $x_{7}$: & \# malignant lesions of M (age $<60$) \\
    $x_{8}$: & \# malignant lesions of M (age $\geq 60$) \\
    $x_{9}$: & \# malignant lesions of F (age $<60$) \\
    $x_{10}$: & \# malignant lesions of F (age $\geq 60$) \\
    $x_{11}$: & \# benign M (age $<60$) \\
    $x_{12}$: & \# benign M (age $\geq 60$) \\
    $x_{13}$: & \# benign F (age $<60$) \\
    $x_{14}$: & \# benign F (age $\geq 60$) \\
\end{tabular}
\end{minipage}
    \subsection{Constraints}
    \setcounter{equation}{0}
{\small\begin{minipage}{\columnwidth}
\begin{align}
    & x_{1}-x_{2}=0 \label{eq:aeq1} \\
    & rx_{4}-x_{3}=0 \label{eq:aeq2} \\
    & sx_{8}-x_{7}=0 \label{eq:aeq3} \\
    & tx_{10}-x_{9}=0 \label{eq:aeq4} \\
    & x_{1}-x_{3}-x_{4}=0 \label{eq:aeq5} \\
    & x_{3}-x_{7}-x_{8}=0 \label{eq:aeq6} \\
    & x_{4}-x_{9}-x_{10}=0 \label{eq:aeq7} \\
    & ux_{12}-x_{11}=0 \label{eq:aeq8} \\
    & vx_{14}-x_{13}=0 \label{eq:aeq9} \\
    & x_{2}-x_{5}-x_{6}=0 \label{eq:aeq10} \\
    & x_{6}-x_{13}-x_{14}=0 \label{eq:aeq11} \\
    & x_{5}-x_{11}-x_{12}=0 \label{eq:aeq12} \\
    & wx_{6}-x_{5}=0 \label{eq:aeq13}
\end{align}

\begin{description}
    \item[\normalfont\textit{Eq.~\ref{eq:aeq1}}:] \# of malignant records equals \# benign records.
    \item[\normalfont\textit{Eq.~\ref{eq:aeq2}}:] Ratio $r$ of malignant male (M) to female patients (F).
    \item[\normalfont\textit{Eq.~\ref{eq:aeq3}}:] Ratio $s$ of malignant M (age $<60$) to M (age $\geq 60$).
    \item[\normalfont\textit{Eq.~\ref{eq:aeq4}}:] Ratio $t$ of malignant F (age $<60$) to F (age $\geq 60$).
    \item[\normalfont\textit{Eq.~\ref{eq:aeq5}}:] \# of malignant lesions equals sum of malignant M and F.
    \item[\normalfont\textit{Eq.~\ref{eq:aeq6}}:] \# M with malignant lesions is equal to that of all ages.
    \item[\normalfont\textit{Eq.~\ref{eq:aeq7}}:] \# F with malignant lesions is equal to that of all ages.
    \item[\normalfont\textit{Eq.~\ref{eq:aeq8}}:] Ratio $u$ of benign M (age $<60$) to M (age $\geq 60$).
    \item[\normalfont\textit{Eq.~\ref{eq:aeq9}}:] Ratio $v$ of benign F (age $<60$) to F (age $\geq 60$).
    \item[\normalfont\textit{Eq.~\ref{eq:aeq10}}:] \# benign lesions equals sum of benign M and F.
    \item[\normalfont\textit{Eq.~\ref{eq:aeq11}}:] \# F with benign lesions is equal to \# all ages.
    \item[\normalfont\textit{Eq.~\ref{eq:aeq12}}:] \# M with benign lesions is equal to \# all ages.
    \item[\normalfont\textit{Eq.~\ref{eq:aeq13}}:] Ratio $w$ of benign M to F.
\end{description}

\end{minipage}

\newpage\section{Age distribution LP model}\label{lp_model_age_dist}
	
    \subsection{Decision variables}
    The age brackets are defined as follows, \rebuttal{where $a$ represents the patient's age in years.}\\
$x_{1}$:\text{\# \rebuttal{${0 \leq a \leq 50} (A_1)$} instances}\\
$x_{2}$:\text{\#  \rebuttal{${51 \leq a \leq 60} (A_2)$} instances}\\
$x_{3}$:\text{\#  \rebuttal{${61 \leq a \leq 70} (A_3)$} instances}\\
$x_{4}$:\text{\#  \rebuttal{${71 \leq a \leq 80} (A_4)$} instances}\\
$x_{5}$:\text{\#  \rebuttal{${a \geq 81} (A_5)$} instances }
        
    \subsection{Objective function}
    \begin{equation*}
        \begin{aligned}
        & \text{Find a vector } x \text{ (decision variables)} \\
        & \text{that maximises } f = x_1 \text{ (objective function)} \\
        & \text{subject to } a_{i1}x_1 + a_{i2}x_2 + \dots + a_{i5}x_5\leq b_i \text{ (constraints)}\\
        & \hspace{2.5em} \text{for } i=1,\dots,5 \\
        & \text{and } x_j\geq 0 \text{ (non-negativity constraints)}\\
        & \hspace{2.5em} \text{for } j=1,\dots,5
        \end{aligned}
    \end{equation*}
    
    \subsection{Constraints}
    \setcounter{equation}{0}
\begin{minipage}{\columnwidth}
    \begin{align}
        & (100-p_1)x_1 - p_1x_2 - p_1x_3 - p_1x_4 - p_1x_5 = 0 \label{eq:aeq001} \\
        & -p_2x_1 + (100-p_2)x_2 - p_2x_3 - p_2x_4 - p_2x_5 = 0 \label{eq:aeq002} \\
        & -p_3x_1 - p_3x_2 + (100-p_3)x_3 - p_3x_4 - p_3x_5 = 0 \label{eq:aeq003} \\
        & -p_4x_1 - p_4x_2 - p_4x_3 + (100-p_4)x_4 - p_4x_5 = 0 \label{eq:aeq004} \\
        & -p_5x_1 - p_5x_2 - p_5x_3 - p_5x_4 + (100-p_5)x_5 = 0 \label{eq:aeq005}
    \end{align}

    \begin{description}
        \item[\normalfont\textit{Eq. \ref{eq:aeq001}}]: Distribution for first category with percentage $p_1$.
        \item[\normalfont\textit{Eq. \ref{eq:aeq002}}]: Distribution for second category with percentage $p_2$.
        \item[\normalfont\textit{Eq. \ref{eq:aeq003}}]: Distribution for third category with percentage $p_3$.
        \item[\normalfont\textit{Eq. \ref{eq:aeq004}}]: Distribution for fourth category with percentage $p_4$.
        \item[\normalfont\textit{Eq. \ref{eq:aeq005}}]: Distribution for fifth category with percentage $p_5$.
    \end{description}
\end{minipage}

\section{PAD-UFES-20 demographics}\label{pad_ufes_lesion_dist}
      \begin{table}[h]
    \caption{Age distribution ($A_1-A_5$) of skin lesions across sex and diagnosis in the curated \textbf{PAD-UFES-20} dataset, showing the breakdown between male and female patients. The age brackets are defined as follows, \rebuttal{where $a$ represents the patient's age in years: $A_1 = {0 \leq a \leq 50}$, $A_2 = {51 \leq a \leq 60}$, $A_3 = {61 \leq a \leq 70}$, $A_4 = {71 \leq a \leq 80}$, and $A_5 = {a \geq 81}$.}}
\centering
\small
\begin{tabular}{lcccc}
Category & Female & Female & Male & Male \\
& benign & malignant & benign & malignant \\
\hline
$A_{1}$ & 34 & 52 & 28 & 56\\
$A_{2}$ & 52 & 110 & 36 & 98 \\
$A_{3}$ & 31 & 88 & 45 & 119 \\
$A_{4}$ & 59 & 88 & 22 & 114 \\
$A_{5}$ & 22 & 63 & 17 & 45\\
\end{tabular}
\label{tab:lesion_stats_padufes}
\end{table}

\end{document}